\documentclass[final,3p,times]{elsarticle}
\usepackage{lineno,hyperref}
\usepackage{amssymb}
\usepackage{amsmath}
\usepackage{amsthm}
\usepackage{mathrsfs}
\usepackage{slashbox}
\usepackage[ruled,vlined]{algorithm2e}
\include{pythonlisting}
\usepackage{setspace}
\usepackage{listings}
\usepackage{tcolorbox}

\usepackage{mathrsfs}
\newtheorem{theorem}{Theorem}[section]

\DeclareMathOperator*{\argminA}{arg\,min}
\modulolinenumbers[200]
\journal{Journal}

\usepackage{subcaption}
\newcommand{\RR}{\mathbb{R}} 
\newcommand{\NN}{\mathbb{N}} 
\newcommand{\hbTheta}{\hat{\boldsymbol{\Theta}}}
\newcommand{\bTheta}{{\boldsymbol{\Theta}}}
\newcommand{\tJ}{\tilde{J}}
\newcommand{\Fcal}{\mathcal{F}}
\newcommand{\T}{^\top}
\DeclareMathOperator{\dom}{dom}
\newcommand{\diag}{\mathop{\mathrm{diag}}}

\begin{document}

\begin{frontmatter}

\title{Locally adaptive activation functions with slope recovery term for deep and physics-informed neural networks}


\author{Ameya D. Jagtap$^{1,*}$, Kenji Kawaguchi$^2$ and George Em Karniadakis$^{1,3}$}
\cortext[mycorrespondingauthor]{Corresponding author Emails:  ameyadjagtap@gmail.com, ameya$\_$jagtap@brown.edu (A.D.Jagtap), kawaguch@mit.edu (K. Kawaguchi), george$\_$karniadakis@brown.edu (G.E.Karniadakis)}

\address{$^1$Division of Applied Mathematics, Brown University, 182 George Street, Providence, RI 02912, USA.
\linebreak $^2$Massachusetts Institute of Technology, 77 Massachusetts Ave, Cambridge, MA 02139, USA.
\linebreak $^3$Pacific Northwest National Laboratory, Richland, WA 99354, USA.}

\begin{abstract}
We propose two approaches of locally adaptive activation functions namely, layer-wise and neuron-wise locally adaptive activation functions, which improve the performance of deep and physics-informed neural networks. The local adaptation of activation function is achieved by introducing a scalable parameter in each layer (layer-wise) and for every neuron (neuron-wise) separately, and then optimizing it using a variant of stochastic gradient descent algorithm. In order to further increase the training speed, an activation slope based \textit{slope recovery} term is added in the loss function, which further accelerates convergence, thereby reducing the training cost. On the theoretical side, we prove that in the proposed method, the gradient descent algorithms are not attracted to sub-optimal critical points or local minima under practical conditions on the initialization and learning rate, and that the gradient dynamics of the proposed method is not achievable by base methods with any (adaptive) learning rates. We further show that the adaptive activation methods accelerate the convergence by implicitly multiplying conditioning matrices to the gradient of the base method without any explicit computation of the conditioning matrix and the matrix-vector product. The different adaptive activation functions are shown to induce different implicit conditioning matrices. Furthermore, the proposed methods with the slope recovery are shown to accelerate the training process. 
\end{abstract}

\begin{keyword}
Machine learning, bad minima, stochastic gradients, accelerated training, PINN, deep learning benchmarks.
\end{keyword}

\end{frontmatter}

\linenumbers

In the recent years, research on neural networks (NNs) has intensified around the world due to their successful applications in many diverse fields such as speech recognition \cite{SpeechRecog}, computer vision \cite{CompVision}, natural language translation \cite{NLT}, \textit{etc}.  
NNs have also been used in the area of scientific computing where they solve partial differential equations (PDEs) due to their ability to effectively approximate complex functions arising in diverse scientific disciplines, cf., Physics-informed Neural Networks (PINNs) by Raissi et al., \cite{RK1} and the references therein. PINNs can accurately solve both direct problems, where the approximate solutions of governing equations are obtained, as well as highly ill-posed inverse problems, where parameters involved in the governing equation are inferred from the training data. 

A major drawback of such NN based models is the slow training/convergence speed, which can adversely affect their performance, especially for solving real-life applications, which require a NN model to run in real-time. Therefore, it is crucial to accelerate the convergence of such models without sacrificing the performance.
Highly efficient and adaptable algorithms are important to design the most effective NN, which not only increase the accuracy of the solution but also reduce the training cost. Various architectures of NN like Dropout NN \cite{DropoutNN} have been proposed in the literature, which can improve the efficiency of the algorithm for specific applications. 
One of the important features of NN is the activation function, which decides the activation of particular neuron during training process. There is no rule of thumb for the choice of activation function, and in fact it solely depends on the problem at hand. Therefore, in this work, we are particularly focusing on adaptive activation functions, which adapt automatically such that the network can be trained faster. Various methods have been proposed in the literature for adaptive activation function, like the adaptive sigmoidal activation function proposed by Yu et al \cite{Yu} for multilayer feedforward NNs. Qian et al \cite{Qian} focused on learning activation functions in convolutional NNs by combining basic activation functions in a data-driven way. Multiple activation functions per neuron were proposed by Dushkoff and Ptucha \cite{MiRa}, where individual neurons select between a multitude of activation functions. Li et al \cite{LLR} proposed a tunable activation function, where only a single hidden layer is used and the activation function is tuned. Shen et al \cite{SWCC} used a similar idea of tunable activation function but with multiple outputs. Recently, Kunc and Kl$\acute{\text{e}}$ma proposed a transformative adaptive activation functions for gene expression inference, see \cite{KK}. One such adaptive activation function was proposed by Jagtap  et al \cite{JK} by introducing a scalable parameter in the activation function, which can be optimized by using any optimization method. Mathematically, it changes the slope of activation function thereby increasing the learning process by altering the loss landscape of NN, especially during the initial training period. Due to single scalar parameter, we call such adaptive activation functions globally adaptive activations, meaning that it gives an optimized slope for the entire network. One can think of doing such optimization at the local level, where the scalable parameters are introduced hidden layer-wise or even for each neuron in the network. Due to different learning capacity of each hidden-layer, such locally defined activation slopes can further improve the performance of the network. In order to further increase the training speed, an activation slope based \textit{slope recovery} term is added in the loss function, which further accelerates convergence.

The rest of the paper is organized as follows. Section 2 presents the methodology of the proposed layer-wise and neuron-wise locally adaptive activations in detail. This also includes a discussion on the slope recovery term, expansion of parametric space due to layer-wise and neuron-wise introduction of additional parameters, and its effect on the overall training cost. In this section, the PINN method is also introduced briefly for completeness. Section 3 gives theoretical results for gradient decent algorithms, where we analyze both the convergent points and gradient dynamics per iteration. In section 4, we perform some computational experiments with the proposed method solving function approximation problem using deep NN and inverse PDE-based problems using PINNs. We also solved some standard deep learning benchmarks problems using the proposed activation functions. Several comparisons are made between the existing and the proposed methods. Finally, in section 5, we summarize the findings of our work.

\section{Methodology}
We use a NN of depth $D$ corresponding to a network with an input layer, $D-1$ hidden-layers and an output layer. In the $k^{th}$ hidden-layer, $N_k$ number of neurons are present. Each hidden-layer of the network receives an output $\mathbf{z}^{k-1} \in\mathbb{R}^{N_{k-1}}$ from the previous layer, where an affine transformation of the form
\begin{equation}\label{afft}
\mathcal{L}_k (\mathbf{z}^{k-1}) \triangleq \mathbf{w}^k \mathbf{z}^{k-1} + \mathbf{b}^k
\end{equation}
is performed. The network weights $\mathbf{w}^k \in\mathbb{R}^{N_k \times N_{k-1}}$ and bias term $\mathbf{b}^k \in\mathbb{R}^{N_{k}}$ associated with the $k^{th}$ layer are chosen from  \textit{independent and identically distributed} sampling. The nonlinear-activation function $\sigma(\cdot)$ is applied to each component of the transformed vector before sending it as an input to the next layer. The activation function is an identity function after an output layer. Thus, the final neural network representation is given by the composition 
$$ u_{\boldsymbol{\Theta}}(\mathbf{z}) = (\mathcal{L}_D \circ \sigma \circ \mathcal{L}_{D-1} \circ \ldots \circ \sigma \circ \mathcal{L}_1 )(\mathbf{z}),$$
where the operator $\circ$ is the composition operator, $\boldsymbol{\Theta} = \{\mathbf{w}^k, \mathbf{b}^k\}_{k=1}^D\in \mathcal{V}$ represents the trainable parameters in the network, and $\mathcal{V}$ is the parameter space; $u$ and $\mathbf{z}^0 = \mathbf{z}$ are the output and input of the network, respectively.

In \cite{JK}, Jagtap et al., proposed an adaptive activation function where an additional scalable parameter $na$, where $n \geq 1$ is pre-defined scaling factor, is introduced. The parameter $a \in \mathbb{R}$ acts as a slope of activation function. Since, the parameter $a$ is defined for the complete network, we are calling it a global adaptive activation function (GAAF). Optimization of such parameter dynamically alters the loss landscape thereby increases the NN convergence, especially during the early training period. It is also possible to extend this strategy by locally defining the activation slope.
In this regard, we propose the following two approaches to locally optimize the activation function.

\vspace{-3mm}
\begin{itemize}
 \item \textbf{Layer-wise locally adaptive activation functions (L-LAAF)}

Instead of globally defining the parameter $a$ for the adaptive activation function, let us define this parameter for each hidden layer as
 $$\sigma(na^k ~\mathcal{L}_k (\mathbf{z}^{k-1})), ~~ k = 1,2,\cdots, D-1.$$
This gives additional $D-1$ parameters to be optimized along with weights and biases. Here, every hidden-layer has its own slope for the activation function.

 \item  \textbf{Neuron-wise locally adaptive activation functions (N-LAAF)}

One can also define such activation function at the neuron level as
 $$\sigma(na_i^k ~(\mathcal{L}_k (z^{k-1}))_i), ~~ k = 1,2,\cdots, D-1,~~ i = 1,2,\cdots, N_k.$$
This gives additional $ \sum_{k=1}^{D-1} N_k$ parameters to be optimized. Neuron-wise activation function acts as a vector activation function in each hidden-layer, where every neuron has its own slope for the activation function, as opposed to scalar activation function given by L-LAAF and GAAF approaches.
\end{itemize}

\vspace{-3mm}
In both cases $n \geq 1$ is a scaling factor. For every problem, there exists a critical scaling factor $n_{crit}$ above which the optimization algorithm becomes very sensitive. The resulting optimization problem leads to finding the minimum of a loss function by optimizing activation slopes along with weights and biases.
Then, the final layer-wise adaptive activation function based neural network representation of the  solution is given by
$$ u_{\hat{\boldsymbol{\Theta}}}(\mathbf{z}) = (\mathcal{L}_D \circ \sigma \circ na^{D-1}\mathcal{L}_{D-1} \circ \sigma \circ na^{D-2}\mathcal{L}_{D-2} \circ \ldots \circ \sigma \circ na^{1}\mathcal{L}_1 )(\mathbf{z}).$$
Similarly, we can write the neuron-wise adaptive activation function based neural network representation of the solution. In this case, the set of trainable parameters $\hat{\boldsymbol{\Theta}}\in \hat{\mathcal{V}}$ consists of $\{\mathbf{w}^k, \mathbf{b}^k\}_{k=1}^{D}$ and $\{a^k_i\}_{k=1}^{D-1}, ~~\forall i = 1,2,\cdots, N_k$. 
In the proposed methods, the initialization of scalable parameters is done such that $na^k_i = 1, ~\forall n$.

The advantage of locally introducing the parameters for the activation slope over its global counterpart is that, it gives additional degrees of freedom to every hidden layer as well as to every neuron in all the hidden layers, which in turn increases the learning capacity of the network. Another advantage of LAAF is that different scaling factors can be assigned for every layer as well as for every neuron as opposed to the global scaling factor in GAAF.

Compared to single additional parameter of global adaptive activation function, the locally adaptive activation function based PINN has several additional scalable parameters to train. Thus, it is important to consider the additional computational cost required.
The increase of the parametric space leads to a high-dimensional optimization problem whose solution can be difficult to obtain.
Between the previously discussed two approaches, \textit{i.e}, L-LAAF and N-LAAF, N-LAAF introduces the highest number of additional parameters for optimization. Next, we discuss the qualitative picture of the increase in the number of parameters. 
Let $\omega$ and $\beta$ be the total number of weights and biases in the NN. Then, the ratio $\mathcal{P}$, which is the size of parametric space of N-LAAF to that of fixed activation based NN is, 
$\mathcal{P} \approx \frac{1+ 2\varrho}{1+\varrho},$
where $\varrho = \beta/\omega$. As an example, consider a fully connected NN with single input and output involving three hidden-layers with 20 neurons in each layer, which gives the values of $\omega = 840$ and $\beta = 61$. Thus, $\mathcal{P} = 1.0677$, i.e., 6.77\% increase in the number of parameters. This increment can be further reduced with an increase in the number of layers as well as neurons in each layer, which eventually results in negligible increase in the number of parameters. In such cases, the computational cost for fixed activation function and that of neuron-wise locally adaptive activations is comparable.

\subsection{Physics-Informed Neural Networks}
In this section we shall briefly introduced the Physics-Informed Neural Network (PINN) algorithm \cite{RK1}.
PINN is a very efficient method for solving forward and inverse differential and integro-differential equations involving noisy, sparse and multi-fidelity data. The main feature of the PINN is that it can easily incorporate all the given information like governing equation, experimental data, initial/boundary conditions etc into the loss function thereby recast the original problem into an optimization problem. One of the main limitation of PINN algorithm is its high computational cost for high-dimensional optimization problem, which is addressed in \cite{JK2} by employing the domain decomposition approach.  The PINN algorithm aims to learn a surrogate $ u = u_{\hat{\boldsymbol{\Theta}}}$ for predicting the solution $u$ of the governing PDE.
In PINN algorithm the loss function is defined as
\begin{equation}\label{loss0}
J(\hat{\boldsymbol{\Theta}}) =  W_{\mathcal{F}}~ MSE_{\mathcal{F}} + W_u ~ MSE_u ,  
\end{equation}
where the mean squared error (MSE) is given as

\begin{align*}
 MSE_{\mathcal{F}}  & = \frac{1}{N_f}\sum_{i=1}^{N_f} |\mathcal{F}_{\hat{\boldsymbol{\Theta}}}(\mathbf{x}^i_f)|^2,
 \\ MSE_u  & =  \frac{1}{N_u}\sum_{i=1}^{N_u} |u^i - u_{\hat{\boldsymbol{\Theta}}}(\mathbf{x}^i_u)|^2.
\end{align*}
$\{\mathbf{x}_f^i\}_{i=1}^{N_f}$ represents the set of residual points, while $\{\mathbf{x}_u^i\}_{i=1}^{N_u}$ represents the training data points. $W_{\mathcal{F}}$ and $W_u$ are the weights for residual and training data points, respectively, which can be chosen dynamically \cite{Wang}.
The neural network solution must satisfy the governing equation given by the residual  $\mathcal{F}_{\hat{\boldsymbol{\Theta}}} = \mathcal{F}(u_{\hat{\boldsymbol{\Theta}}})$ evaluated at randomly chosen residual points in the domain. 
As an example, for the general differential equation of the form $L u = f$, the residual term is given by $$\mathcal{F}_{\hat{\boldsymbol{\Theta}}} \triangleq Lu_{\hat{\boldsymbol{\Theta}}} - f, $$ where $L$ represent the linear/nonlinear differential terms. To construct the residuals in the loss function, derivatives of the solution with respect to the independent variables are required, which can be computed using the \textit{automatic differentiation} (AD) \cite{AD}. AD is an accurate way to calculate derivatives in a computational graph compared to numerical differentiation since they do not suffer from the errors such as truncation and round-off errors. Thus, the PINN method is a grid-free method, which does not require mesh for solving equations.
This constitutes the physics-informed part of neural network as given by the first term in equation \eqref{loss0}.
The second term in equation \eqref{loss0} includes the known boundary/initial conditions, experimental data, which must be satisfied by the neural network solution. 

The resulting optimization problem leads to finding the minimum of a loss function by optimizing the trainable parameters $ \hat{\boldsymbol{\Theta}} $.
The solution to this minimization problem can be approximated iteratively by one of the forms of gradient descent algorithm. The stochastic gradient descent (SGD) algorithm is widely used in machine learning community see, \cite{Rud} for a complete survey. In this work, the ADAM optimizer \cite{ADAM}, which is a variant of the SGD method is used.

\subsection{Loss function with slope recovery term}
The main motivation of adaptive activation function is to increase the slope of activation function, resulting in non-vanishing gradients and fast training  of the network. It is clear that one should quickly increase the slope of activation in order to improve the performance of NN. Thus, instead of only depending on the optimization methods, another way to achieve this is to include the slope recovery term $\mathcal{S}(a) $ defined as
\begin{equation*}
\mathcal{S}(a)  \triangleq \begin{cases} \frac{1}{ \frac{1}{D-1}\sum_{k=1}^{D-1} \text{exp}(a^k)} & \text{for L-LAAF}, \\ \frac{1}{\frac{1}{D-1}\sum_{k=1}^{D-1}\text{exp}\left(\frac{\sum_{i=1}^{N_k} a_i^k}{N_k}\right)} & \text{for N-LAAF}.
                  \end{cases}
\end{equation*}
The main reason behind this is that such term contributes to the gradient of the loss function without vanishing.
The overall effect of including this term is that it forces the network to increase the value of activation slope quickly thereby increasing the training speed. 

\begin{figure} [htpb] 
\centering 
\includegraphics[trim=0cm 2.2cm 2.4cm 1cm, clip=true, scale=0.55, angle = 0]{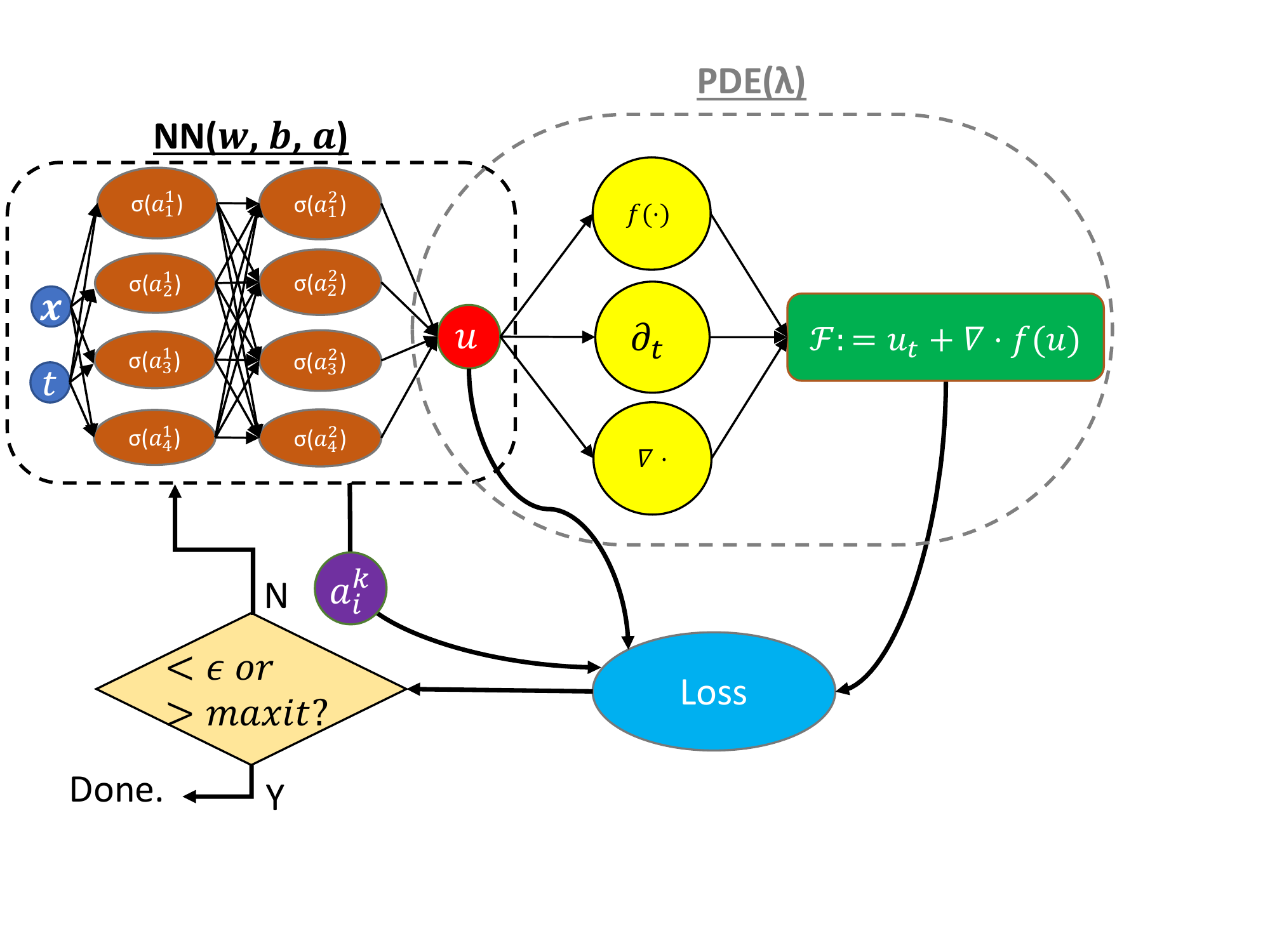}
\caption{Schematic of LAAF-PINN for the Burgers equation. The left NN is the uninformed network while the right one induced by the governing differential equation is the informed network. The input are the coordinates (x,t) while the output is the solution u(x,t) which has to satisfy the governing equation. The two NNs share parameters and they both contribute to the loss function.}
\label{fig:LAAFPINN}
\end{figure}

Figure \ref{fig:LAAFPINN} shows a sketch of a neuron-wise \textit{locally adaptive activation function based physics-informed neural network} (LAAF-PINN), where both the NN part along with the physics-informed part can be seen. 
The activation slopes from every neuron are also contributing to the loss function in the form of slope recovery term. 
The following algorithm summarizes the LAAF-PINN algorithm with slope recovery term.

\begin{algorithm}[H]
\SetAlgoLined
 \textbf{Step 1} : Specification of training set in computational domain \\
\textit{Training data} : $u_{\hat{\boldsymbol{\Theta}}}$ network $\{\mathbf{x}_u^i\}_{i=1}^{N_u}$, ~~ \textit{Residual training points} :   $\mathcal{F}_{\hat{\boldsymbol{\Theta}}}$ network $\{\mathbf{x}_f^i\}_{i=1}^{N_f}$ \\
 \textbf{Step 2} : Construct neural network $u_{\hat{\boldsymbol{\Theta}}}$ with random initialization of parameters $\hat{\boldsymbol{\Theta}}$. \\
  \textbf{Step 3} : Construct the residual neural network  $\mathcal{F}_{\hat{\boldsymbol{\Theta}}}$ by
  substituting surrogate $u_{\hat{\boldsymbol{\Theta}}}$ into the governing equations using 
automatic differentiation and other arithmetic operations.\\
\textbf{Step 4} : Specification of the loss function that includes the slope recovery term:\\
\begin{equation}\label{loss}
\tilde{J}(\hat{\boldsymbol{\Theta}}) = \frac{W_{\mathcal{F}}}{N_f}\sum_{i=1}^{N_f} |\mathcal{F}_{\hat{\boldsymbol{\Theta}}}(\mathbf{x}^i_f)|^2+ \frac{W_u}{N_u}\sum_{i=1}^{N_u} |u^i - u_{\hat{\boldsymbol{\Theta}}}(\mathbf{x}^i_u)|^2 + W_a ~\mathcal{S}(a),
\end{equation}\\
where $W_a$ is the weight for slope recovery term.\\
\textbf{Step 5} : Find the best parameters $ \hat{\boldsymbol{\Theta}}^*$ using a suitable optimization method for minimizing the loss function $\tilde{J}(\hat{\boldsymbol{\Theta}})$ as
$$\hat{\boldsymbol{\Theta}}^* = \argminA_{\hat{\boldsymbol{\Theta}} \in  \hat{\mathcal{V}}} ~\tilde{J}(\hat{\boldsymbol{\Theta}}).$$
 \caption{LAAF-PINN algorithm with slope recovery term.}
\end{algorithm}

\section{Gradient dynamics with adaptive activation: convergent points and acceleration of convergence} \label{sec:theory}

When compared with the standard method, the adaptive activation method induces a new gradient dynamics, which results in different convergent points and the acceleration of the convergence. The following theorem states that a gradient descent algorithm minimizing our objective function $\tilde J(\hbTheta)$  in \eqref{loss} does not converge to a sub-optimal critical point or a sub-optimal local minimum,  for neither L-LAAF nor N-LAAF,  given appropriate initialization and learning rates. For simplicity, $W_{\mathcal{F}}, W_u$ and $W_a$ are assumed to be unity. In the following theorem, we treat $\hbTheta$ as a real-valued vector.
Let $\tJ c(0)=MSE_{\mathcal{F}}+MSE_u$ with the constant network $u_{\bTheta}(z)=u_{\bTheta}(z')=c \in \RR^{N_D}$ for all $z,z'$ where $c$ is a constant.  
\begin{theorem} \label{thm1}
Let $(\hbTheta_m)_{m\in \NN}$ be a sequence generated by a gradient descent algorithm as $\hbTheta_{m+1}=\hbTheta_{m}-\eta_m\nabla \tJ(\hbTheta)$. Assume that $\tJ(\hbTheta_0)<\tJ c(0)+\mathcal{S}(0)$ for any $c \in \RR^{N_D}$, $\tJ$  is differentiable, and that for each $i\in \{1,\dots,N_f\}$, there exist differentiable function $\varphi^{i}$ and input $\rho^{i}$ such that  $|\Fcal_{\hat{\boldsymbol{\Theta}}}(\mathbf{x}^i_f)|^2=\varphi^{i}(u_{\hbTheta}(\rho^i) )$. Assume that at least one of the following three conditions holds.  
\begin{description}
\item[(i)] 
(constant learning rate)   $\nabla \tJ$ is Lipschitz continuous with Lipschitz constant $C$ (i.e.,  $\|\nabla \tJ(\hbTheta)-\nabla \tJ(\hbTheta')\|_2 \le C \|\hbTheta - \hbTheta'\|_2$ for all $\hbTheta,\hbTheta'$ in its domain), and  $\epsilon \le \eta_m \le (2-\epsilon)/C$, where $\epsilon$ is a fixed positive number. \item[(ii)]
(diminishing  learning rate)  $\nabla \tJ$ is Lipschitz continuous,  $\eta_m \rightarrow 0$ and $\sum_{m=0}^\infty \eta_m =\infty$. \item[(iii)] 
(adaptive learning rate) the learning rate  $\eta_m$ is chosen by the minimization rule, the limited minimization rule, the Armjio rule, or the Goldstein rule \cite{bertsekas1999nonlinear}. 
\end{description}
Then, for both L-LAAF and N-LAAF, no limit point of   $(\hbTheta_m)_{m\in \NN}$ is   a sub-optimal critical point or  a sub-optimal local
minimum.

\end{theorem}
The initial condition $\tJ(\hbTheta_0)<\tJ c(0)+\mathcal{S}(0)$ means that the initial value $\tJ(\hbTheta_0)$ needs to be less than that of  a constant network plus the highest value of the slope recovery term. Here, note that $\mathcal{S}(1)<\mathcal{S}(0)$.  The proof of Theorem \ref{thm1} is included in the appendix.

We now study how the proposed method approaches the convergent points and why it can accelerate the convergence. To illustrate the principle mechanism behind the acceleration,  we compare the gradient dynamics of the  proposed method against that of the standard method in the domain of $J$.  The gradient dynamics of the standard method in the domain of $J$ is
\begin{equation} \label{eq:gradient_dynamics_standard}
\bTheta^{m+1} = \bTheta^{m} - \eta _{m}\nabla J(\bTheta^{m}), 
\end{equation}
and generates the sequence  $(J(\bTheta^{m} ))_{m\in \NN}$ of the standard objective values. The gradient dynamics of  the adaptive activation method in the domain of $\tilde J$ is  $\hbTheta^{m+1} = \hbTheta^{m} - \eta _{m}\nabla \tilde J(\hbTheta^{m})$, and   generates the sequence  $(\tilde J(\hbTheta^{m} ))_{m\in \NN}$ of the modified  objective values. Those dynamics are in the two different spaces, the domains of $J$  and  $\tilde J$. To compare them, we translate the  dynamics $(\hbTheta^{m} )_{m\in \NN}$ in the domain of $\tilde J$  to the dynamics $(\tilde \bTheta^{m} )_{m\in \NN}$ in the domain of $J$. 

More concretely, we show  that the gradient dynamics  $(\hbTheta^{m} )_{m\in \NN}$ of the \textit{global} adaptive activation method generates the sequence  $(J(\tilde \bTheta^{m} ))_{m\in \NN}$ of the standard objective values where
\begin{align} \label{eq:gradient_dynamics_gaaf}
\tilde \bTheta^{m+1} 
=\tilde\bTheta^{m} -\eta_{m} \hat G (\hat \bTheta^{m+1})\nabla J(\tilde\bTheta^{m})+\eta_{m}^2\hat H_{J} (\tilde\bTheta^{m})\tilde\bTheta^{m},
\end{align}
where $\tilde \bTheta^{m+1} \in \dom(J)$ is $\hat \bTheta^{m+1}\in \dom(\tilde J)$ being translated in the space of $\bTheta^{m+1}$, which is $\dom(J)$ ($\dom(J)$ is the domain of $J$),    
$$
\hat G (\hat \bTheta^{m+1}) = (a^{m})^2  I+W^{m} (W^{m})^\top,
$$
and
$$
\hat H_{J} (\tilde\bTheta^{m})=\nabla J(\tilde\bTheta^{m})\nabla J(\tilde\bTheta^{m})^\top.
$$
Comparing equations \eqref{eq:gradient_dynamics_standard} and \eqref{eq:gradient_dynamics_gaaf}, we can see that the gradient dynamics of the adaptive activation modifies the standard dynamics, by multiplying a conditioning matrix $\hat G (\tilde\bTheta^{m})$ to the gradient and by adding the approximate second-order term  $\eta_{m}^2H (\tilde\bTheta^{m})\tilde\bTheta^{m}$. This  provides the mathematical intuition of why the global adaptive activation method can accelerate the convergence and it is not equivalent to changing or adapting learning rates.

To understand the approximate second-order term   $\eta_{m}^2H (\tilde\bTheta^{m})\tilde\bTheta^{m}$, note that  the gradient dynamics of the standard method \eqref{eq:gradient_dynamics_standard} can be viewed as the simplest discretization (Euler's Method) of the    differential equation of the gradient flow, 
\begin{equation} \label{eq:gradient_flow}
\dot \bTheta = -\nabla J(\bTheta),
\end{equation}
where we  approximate $\bTheta^{t_{0}+t_{1}} =\bTheta^{t_{0}} - \int_{t\in[t_0,t_1]}\nabla J(\bTheta^{t})dt$ by setting $\nabla J(\bTheta^{t})\approx\nabla J(\bTheta^{t_0})$ for $t\in[t_0,t_1]$. In this view, 
we can consider other discretization  procedures of \eqref{eq:gradient_flow}. For example, instead of setting $\nabla J(\bTheta^{t})\approx\nabla J(\bTheta^{t_0})$, we can approximate 
$$
\nabla J(\bTheta^{t})\approx\nabla J(\bTheta^{t_0})+H_{J}(\bTheta^{t_0})(\bTheta^{t}-\bTheta^{t_0}),
$$
where $H_{J}$
is the Hessian of $J$. The term    $\eta_{m}^2H (\tilde\bTheta^{m})\tilde\bTheta^{m}$ in  \eqref{eq:gradient_dynamics_gaaf} can be obtained by further approximating the second term by setting $H_{J}(\bTheta^{t_0})\approx\hat H_{J} (\tilde\bTheta^{t_0})$ and $\bTheta^{t}-\bTheta^{t_0}\approx-\eta_{m}\tilde\bTheta^{t_0}$.

More generally, we show  that the gradient dynamics  $(\hbTheta^{m} )_{m\in \NN}$ of \textit{any}  adaptive activation method generates the sequence  $(J(\tilde \bTheta^{m} ))_{m\in \NN}$ of the standard objective values, where
\begin{align} \label{eq:gradient_dynamics_general}
\tilde \bTheta^{m+1} 
=\tilde\bTheta^{m} -\eta_{m} G (\hat \bTheta^{m+1})\nabla J(\tilde\bTheta^{m}),
\end{align}
with 
$$
G(\hat\bTheta^{m+1}) =\diag((Aa^{m})^2)+\diag(W^{m} )AA^{\T}    \diag(W^{m} )-\eta_{m} \diag(V(\hat\bTheta^{m+1})),
$$
and
$$
V(\hat\bTheta^{m+1}) =\diag(Aa^{m}  )AA^{\T}    \diag(W^{m} )\nabla J(\tilde\bTheta^{m}).
$$
Here, given a vector $v \in \RR^d$, $\diag(v)\in \RR^{d\times d}$ represents a diagonal matrix with $\diag(v)_{ii}=v_i$ and $v^{2}$ represents $v \circ v$ where    $v\circ u$  is the element-wise product of the two vectors $v$ and $u$. The matrix $A$ differs for different methods of adaptive activation functions with different types of locality, and is  a fixed matrix given a method of  GAAF, L-LAAF or N-LAAF.
 For example, in the case of GAAF, $d'=1$ and $A=(1,1,\dots,1)\T \in \RR^d$.
By plugging this $A$ into \eqref{eq:gradient_dynamics_general} and noticing that $\diag(W^{m} )AA^{\T}    \diag(W^{m} )=WW^{\T}$ with this $A$, we can  obtain   \eqref{eq:gradient_dynamics_gaaf} from  \eqref{eq:gradient_dynamics_general}; i.e.,  \eqref{eq:gradient_dynamics_general} is a strictly more general version of \eqref{eq:gradient_dynamics_gaaf}.
In the general case for any  adaptive activation method with different types of locality,  we can write $\tilde\bTheta^{m+1} = Aa^{m+1} \circ W^{m+1}$, where   $W^{m+1} \in \RR^d$ and $a^{m+1} \in \RR^{d'}$ with some matrix $A\in \RR^{d \times d'}$. In the case of L-LAAF, $d'$ is the number of layers and $A\in \RR^{d \times d'}$ is the matrix that satisfies $\tilde\bTheta^{m+1} = Aa^{m+1} \circ W^{m+1}$ for L-LAAF. In the case of N-LAAF, $d'$ is the number of all neurons and $A\in \RR^{d \times d'}$ is the matrix that satisfies $\tilde\bTheta^{m+1} = Aa^{m+1} \circ W^{m+1}$ for N-LAAF. 

Comparing equations \eqref{eq:gradient_dynamics_standard} and \eqref{eq:gradient_dynamics_general}, we can see that the gradient dynamics of different  adaptive activation methods modify the standard dynamics, by multiplying different conditioning matrices $G (\hat\bTheta^{m})$ to the gradient, with different matrices $A$. This  provides the mathematical intuition of why various adaptive activation methods can accelerate the convergence in different ways with different matrices $A$, and they are not equivalent to changing or adapting learning rates.
Also, our analysis with   \eqref{eq:gradient_dynamics_general} is applicable to any adaptive activation methods beyond GAAF, L-LAAF and N-LAAF, and provides insights for designing new adaptive activation methods that correspond to the new matrices $A$ in \eqref{eq:gradient_dynamics_general} in order to further accelerate the convergence. 

The conditioning matrix $G (\hat\bTheta^{m})$ is positive definite with sufficiently small learning rate $\eta_m$ since $\diag((Aa^{m})^2)$ is positive definite and $\diag(W^{m} )AA^{\T}    \diag(W^{m} )$ is positive semi-definite (when every entry of $a^{m}$ is nonzero).
Therefore, with sufficiently small learning rate $\eta_m$, the parameter update in equation \eqref{eq:gradient_dynamics_general} decreases the value of $J$ as $J(\bTheta^{m+1}) < J(\bTheta^{m})$ at differentiable points whenever $\|\nabla J(\tilde\bTheta^{m})\| \neq 0$. This is because $J(\bTheta^{m+1}) =J(\bTheta^{m})-\eta_{m} \nabla J(\tilde\bTheta^{m})^\top G (\hat \bTheta^{m+1})\nabla J(\tilde\bTheta^{m}) + \eta_{m} \varphi( \eta_{m} )$ at differentiable points with some function $\varphi$ such that $\lim_{\eta_m \rightarrow 0}\varphi(\eta_m)=0$. 
 
We now derive equation \eqref{eq:gradient_dynamics_gaaf} and  \eqref{eq:gradient_dynamics_general}, and explain the definition of each symbol in more detail. 
Let us first focus on the GAAF method without the recovery term. Let  $\hat J$ be $\tilde J$ without the recovery term. Let $\tilde \bTheta=aW$, where $W$ consists of all standard weight and bias parameters $\{\mathbf{w}^k, \mathbf{b}^k\}_{k=1}^{D}$. Then, we have that $\hat J(\hbTheta)=J(\tilde \bTheta)$. While   the standard method generates the sequence of standard objective values $(J(\bTheta^{m+1} ))_{m\in \NN}$, the GAAF method generates the sequence of standard objective values $(J(\tilde \bTheta^{m}))_{m\in \NN}$. To compare the two methods in terms of the same standard objective values, we are thus interested in the following gradient dynamics of the proposed method:
\begin{align} \label{eq:derivation}
\tilde \bTheta^{m+1} &= a^{m+1} W^{m+1} 
\\ \nonumber & =(a^{m}  - \eta_{m} \nabla_{a} J(\tilde \bTheta^{m}))(W^{m} - \eta_{m} \nabla_{W} J(\tilde \bTheta^{m}))
\\ \nonumber&=\tilde\bTheta^{m} -\eta_{m} a^{m} \nabla_{W} J(\tilde \bTheta^{m})-\eta_{m} \nabla_{a} J(\tilde \bTheta^{m})W^{m} +\eta_{m}^2 \nabla_{a} J(\tilde \bTheta^{m})\nabla_{W} J(\tilde \bTheta^{m})   
\end{align}
Here, we have that  
$
\nabla_{W} J(\tilde \bTheta) = \left( \frac{\partial J(\tilde \bTheta)}{\partial\tilde \bTheta}\frac{\partial \tilde \bTheta}{ \partial W} \right)^\top
$
and
$
\nabla_{a} J(\tilde \bTheta) = \left( \frac{\partial J(\tilde \bTheta)}{\partial\tilde \bTheta}\frac{\partial \tilde \bTheta}{ \partial a} \right)^\top$ by the chain rule, where  $\frac{\partial J(\tilde \bTheta)}{\partial\tilde \bTheta}= \nabla J(\tilde \bTheta)^\top$, $\frac{\partial \tilde \bTheta}{ \partial W}=aI$ ($I$ is the identity matrix) and $\frac{\partial \tilde \bTheta}{ \partial a}=W$. By plugging these into each term in the last line of \eqref{eq:derivation}, $$
a^{m} \nabla_{W} J(\tilde \bTheta^{m}) =(a^{m})^2  \nabla J(\tilde\bTheta^{m}),
$$
$$
\nabla_{a} J(\tilde \bTheta^{m})W^{m}=W^{m} (W^{m})^\top\nabla J(\tilde\bTheta^{m}), $$
and 
$$
\nabla_{a} J(\tilde \bTheta^{m})\nabla_{W} J(\tilde \bTheta^{m})=\nabla J(\tilde\bTheta^{m})\nabla J(\tilde\bTheta^{m})^\top\tilde\bTheta^{m},
$$
which obtains  \eqref{eq:gradient_dynamics_gaaf} as desired. Similarly to the case of GAAF, in the general case,
we have\begin{align} \label{eq:derivation2}
\tilde \bTheta^{m+1} &=Aa^{m+1} \circ W^{m+1}
\\ \nonumber & =(A(a^{m}  - \eta_{m} \nabla_{a} J(\tilde \bTheta^{m}))) \circ(W^{m} - \eta_{m} \nabla_{W} J(\tilde \bTheta^{m}))
\\ \nonumber&=\scalebox{0.95}{$\displaystyle \tilde\bTheta^{m} -\eta_{m} (Aa^{m} \circ \nabla_{W} J(\tilde \bTheta^{m}))-\eta_{m} (A\nabla_{a} J(\tilde \bTheta^{m})\circ W^{m} )+\eta_{m}^2 (A\nabla_{a} J(\tilde \bTheta^{m})\circ \nabla_{W} J(\tilde \bTheta^{m})).$} \end{align}  
By using  
$
\nabla_{W} J(\tilde \bTheta) = \left( \frac{\partial J(\tilde \bTheta)}{\partial\tilde \bTheta}\frac{\partial \tilde \bTheta}{ \partial W} \right)^\top
$
and
$
\nabla_{a} J(\tilde \bTheta) = \left( \frac{\partial J(\tilde \bTheta)}{\partial\tilde \bTheta}\frac{\partial \tilde \bTheta}{ \partial a} \right)^\top$, 
\begin{align}
Aa^{m} \circ \nabla_{W} J(\tilde \bTheta^{m})= \diag(Aa^{m} \circ Aa^{m})   \nabla J(\tilde\bTheta^{m}), 
\end{align}  
\begin{align}
A\nabla_{a} J(\tilde \bTheta^{m})\circ W^{m} = \diag(W^{m} )AA^{\T}    \diag(W^{m} )\nabla J(\tilde\bTheta^{m}), 
\end{align}  
and 
\begin{align}
A\nabla_{a} J(\tilde \bTheta^{m})\circ \nabla_{W} J(\tilde \bTheta^{m})=\diag(\diag(Aa^{m}  )AA^{\T}    \diag(W^{m} )\nabla J(\tilde\bTheta^{m}))\nabla J(\tilde\bTheta^{m}).
\end{align} 
By plugging these into \eqref{eq:derivation2}, we obtain \eqref{eq:gradient_dynamics_general}.

\begin{figure}[ht!]
\begin{subfigure}[b]{0.32\columnwidth}
  \includegraphics[width=\columnwidth,height=0.6\columnwidth]{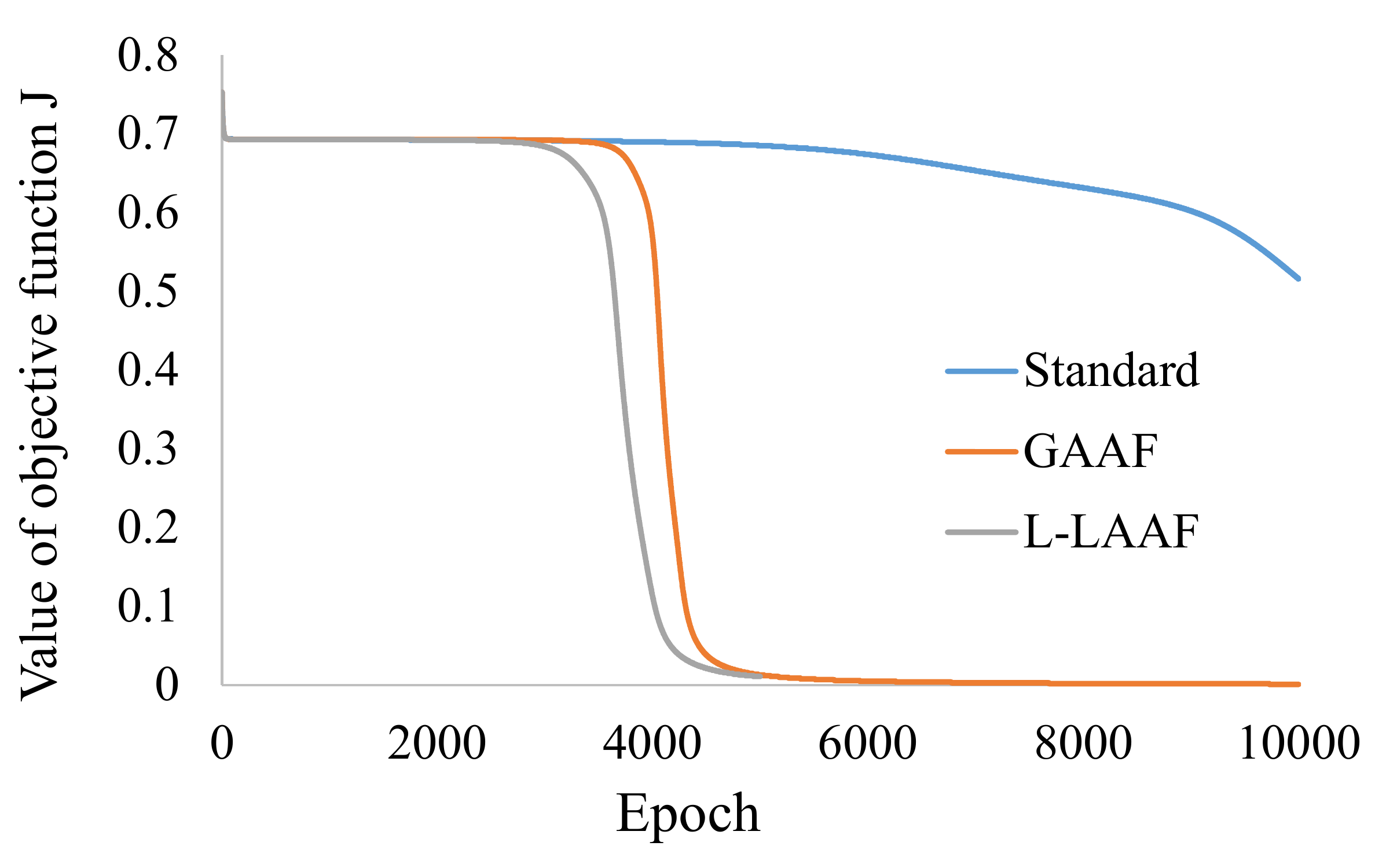}
  \includegraphics[width=\columnwidth,height=0.6\columnwidth]{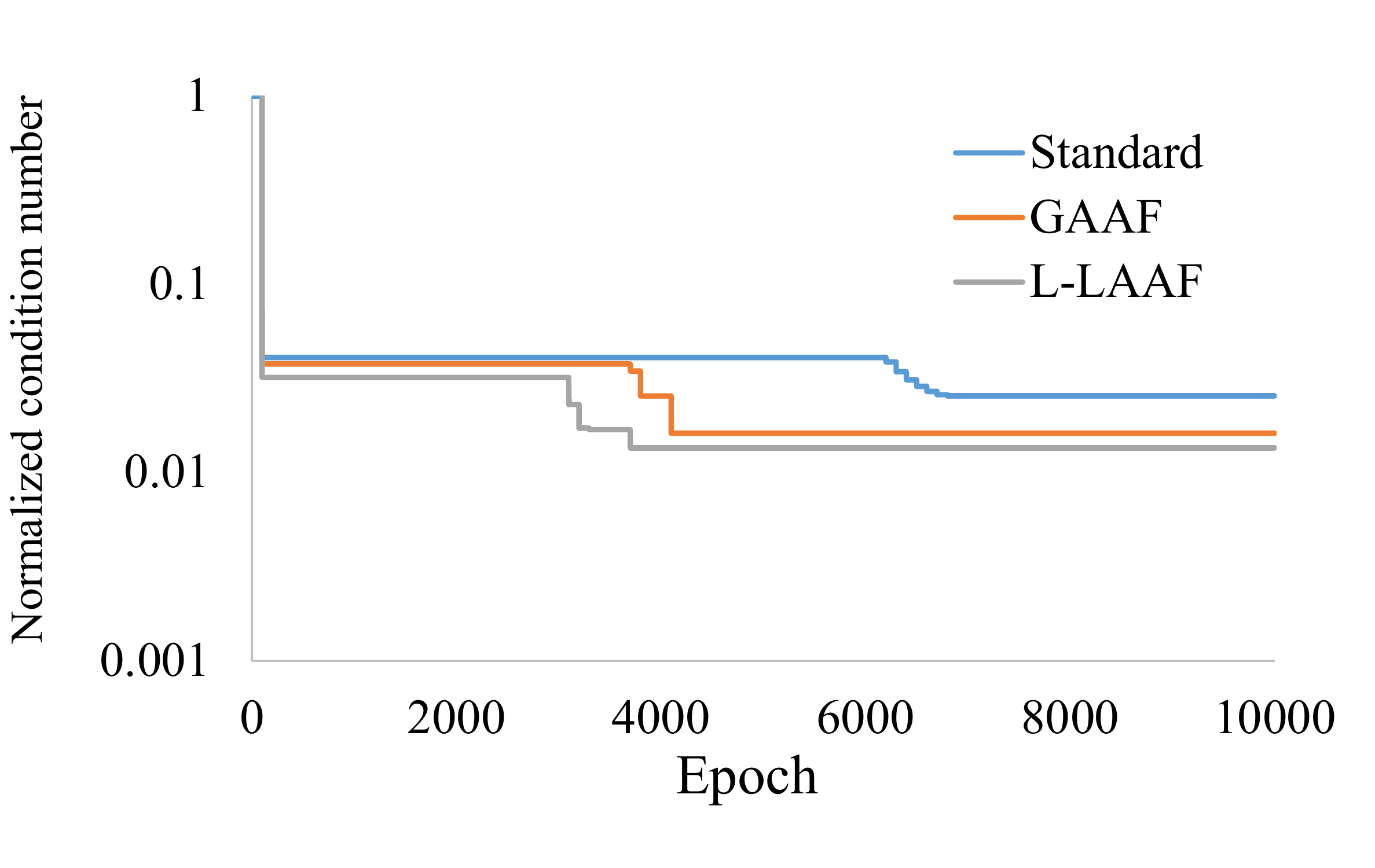}
  \vspace{-20pt}
  \caption{(Sigmoid, width=10)}
  \vspace{10pt}
\end{subfigure}
\begin{subfigure}[b]{0.32\columnwidth}
  \includegraphics[width=\columnwidth,height=0.6\columnwidth]{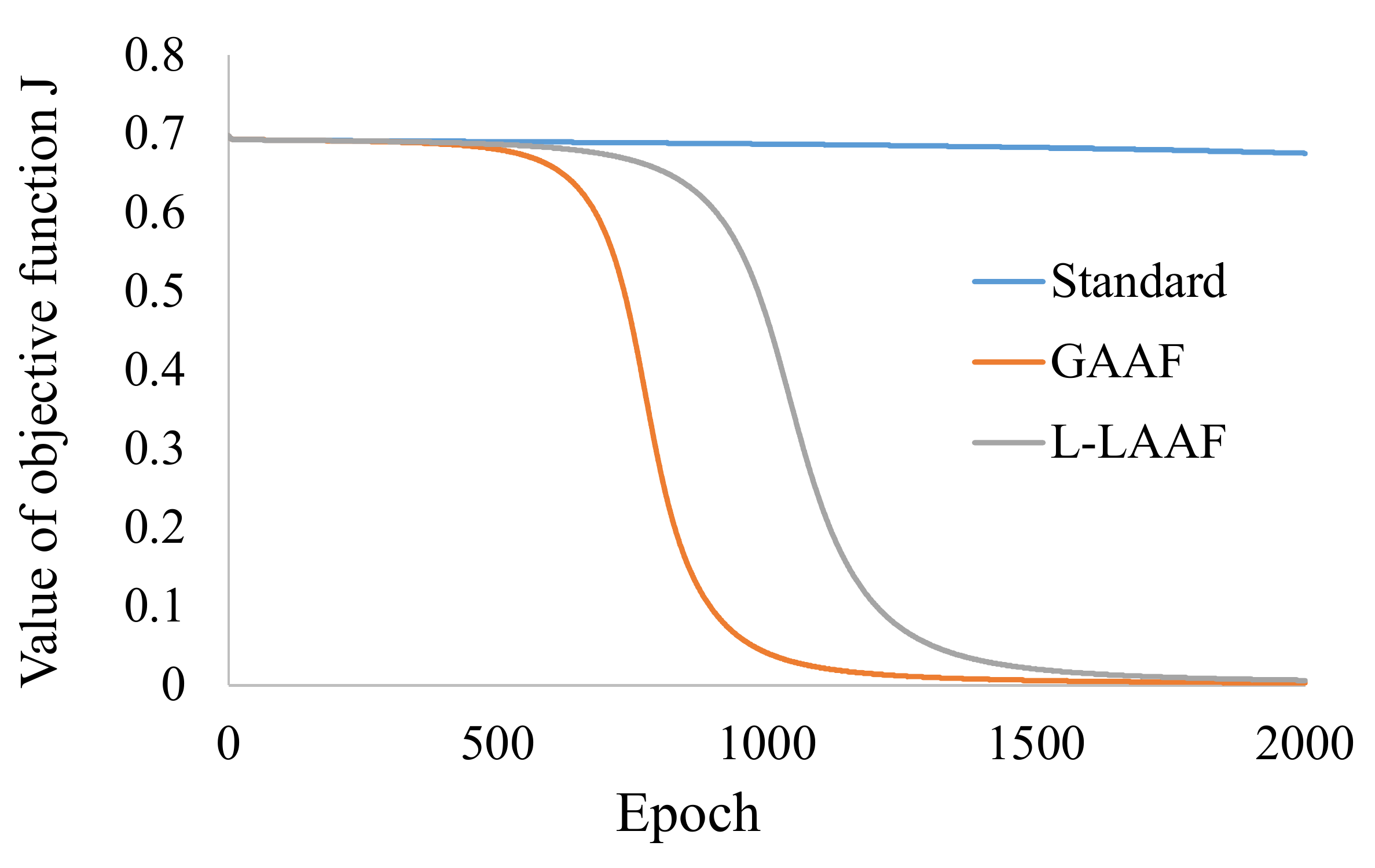}
  \includegraphics[width=\columnwidth,height=0.6\columnwidth]{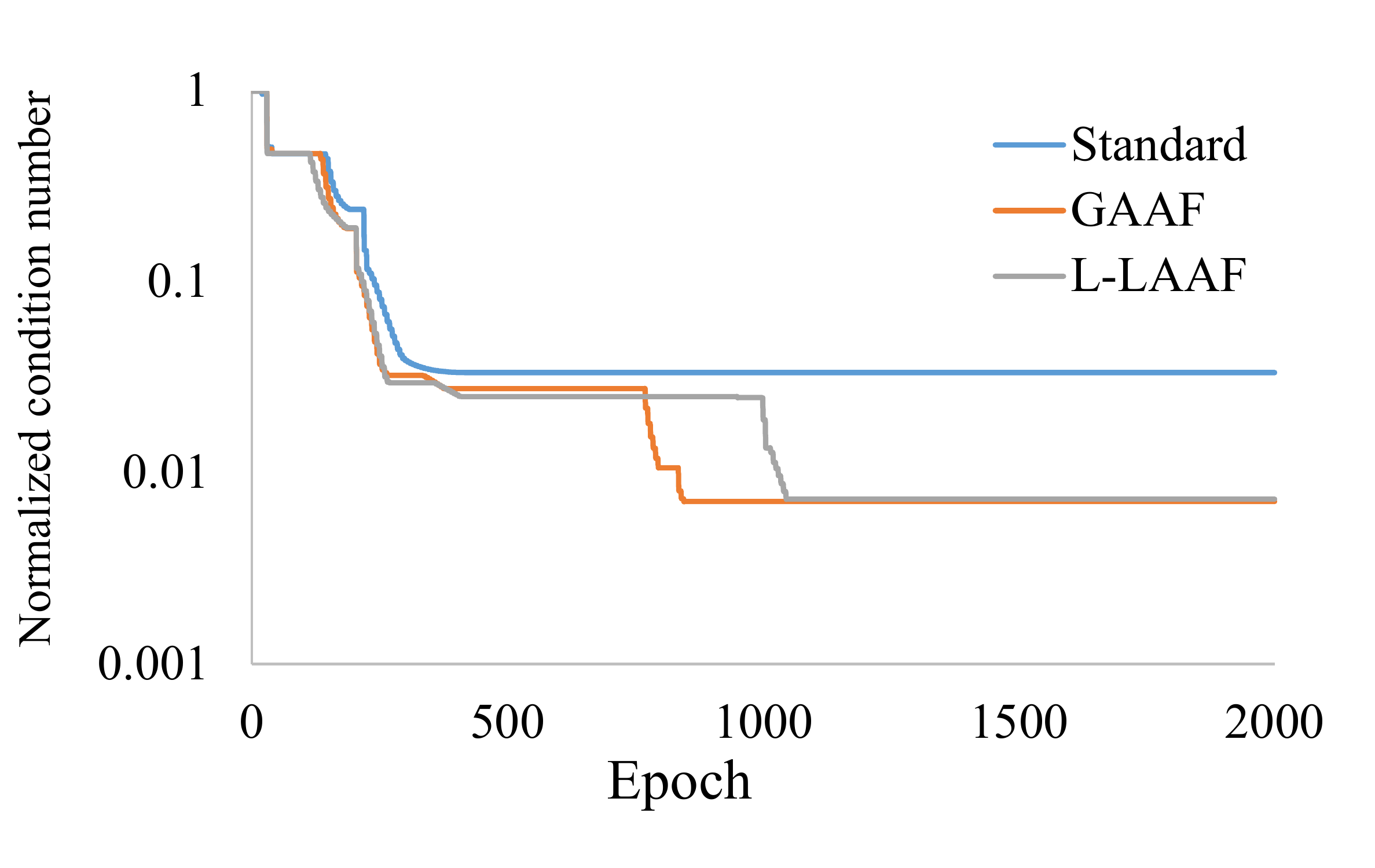}
  \vspace{-20pt}
  \caption{(Sigmoid, width=20)}
  \vspace{10pt} 
\end{subfigure}
\begin{subfigure}[b]{0.32\columnwidth}
  \includegraphics[width=\columnwidth,height=0.6\columnwidth]{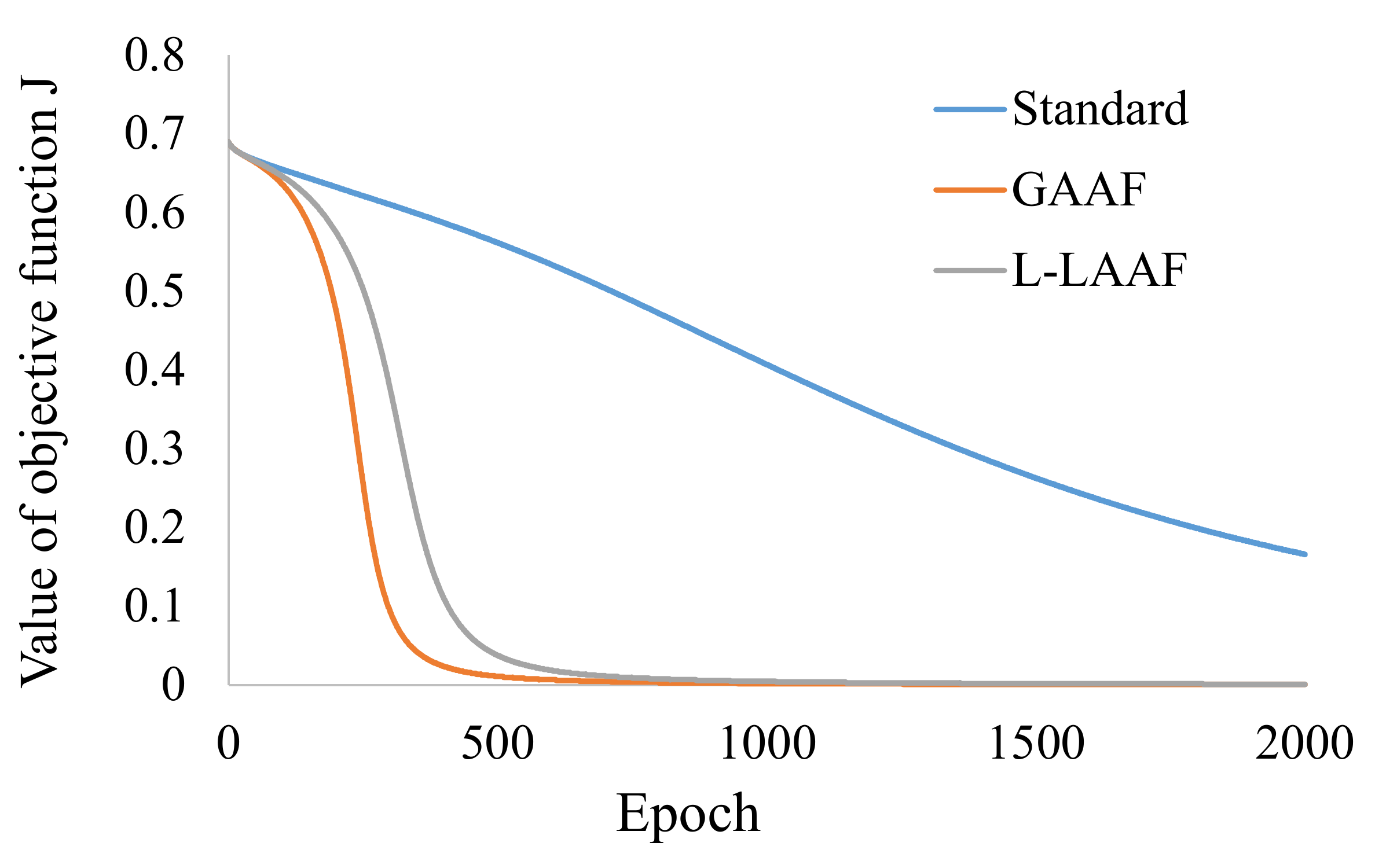}
  \includegraphics[width=\columnwidth,height=0.6\columnwidth]{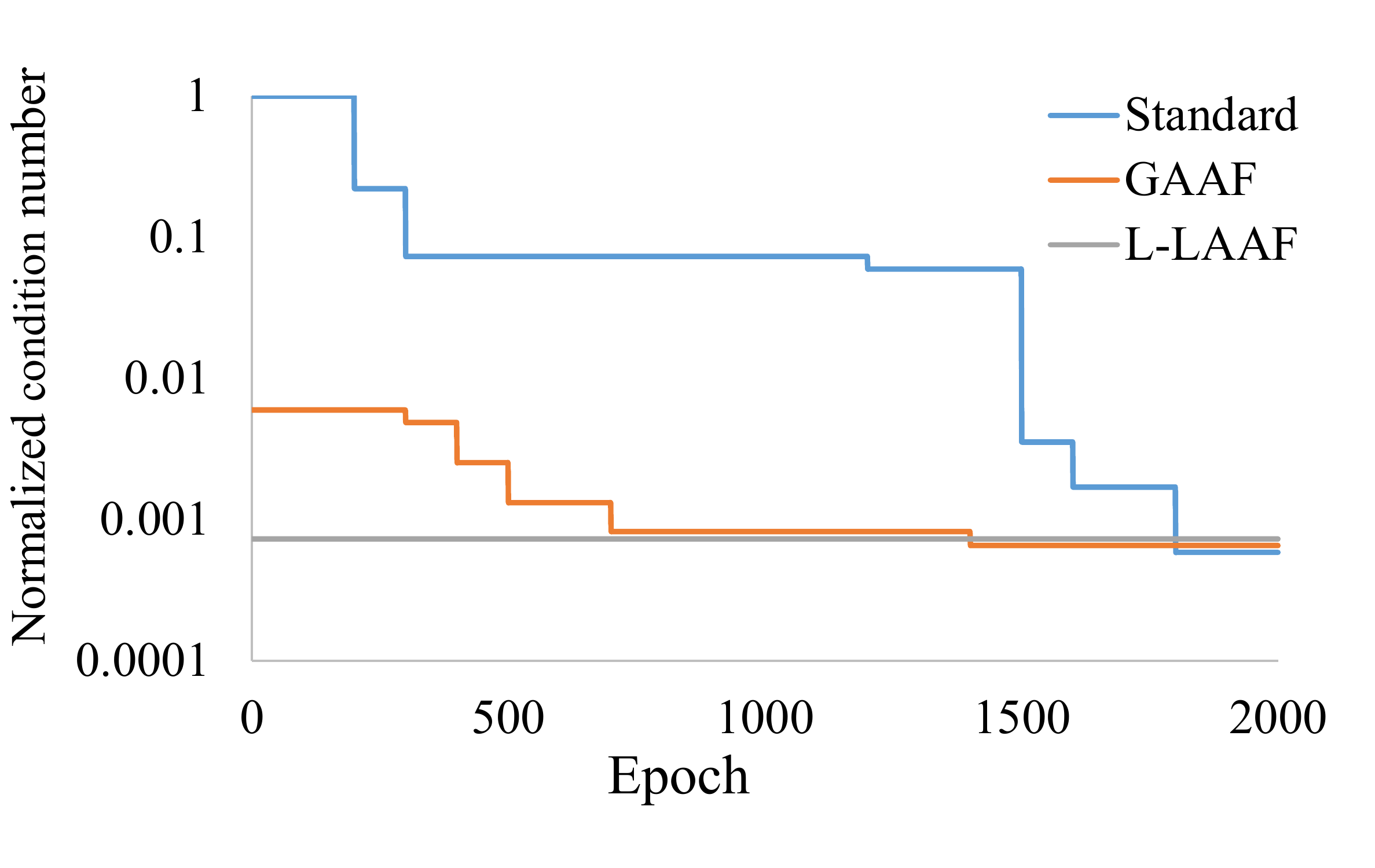}
  \vspace{-20pt}
  \caption{(ReLU, width=20)}
  \vspace{10pt}
\end{subfigure}
\begin{subfigure}[b]{0.32\columnwidth}
  \includegraphics[width=\columnwidth,height=0.6\columnwidth]{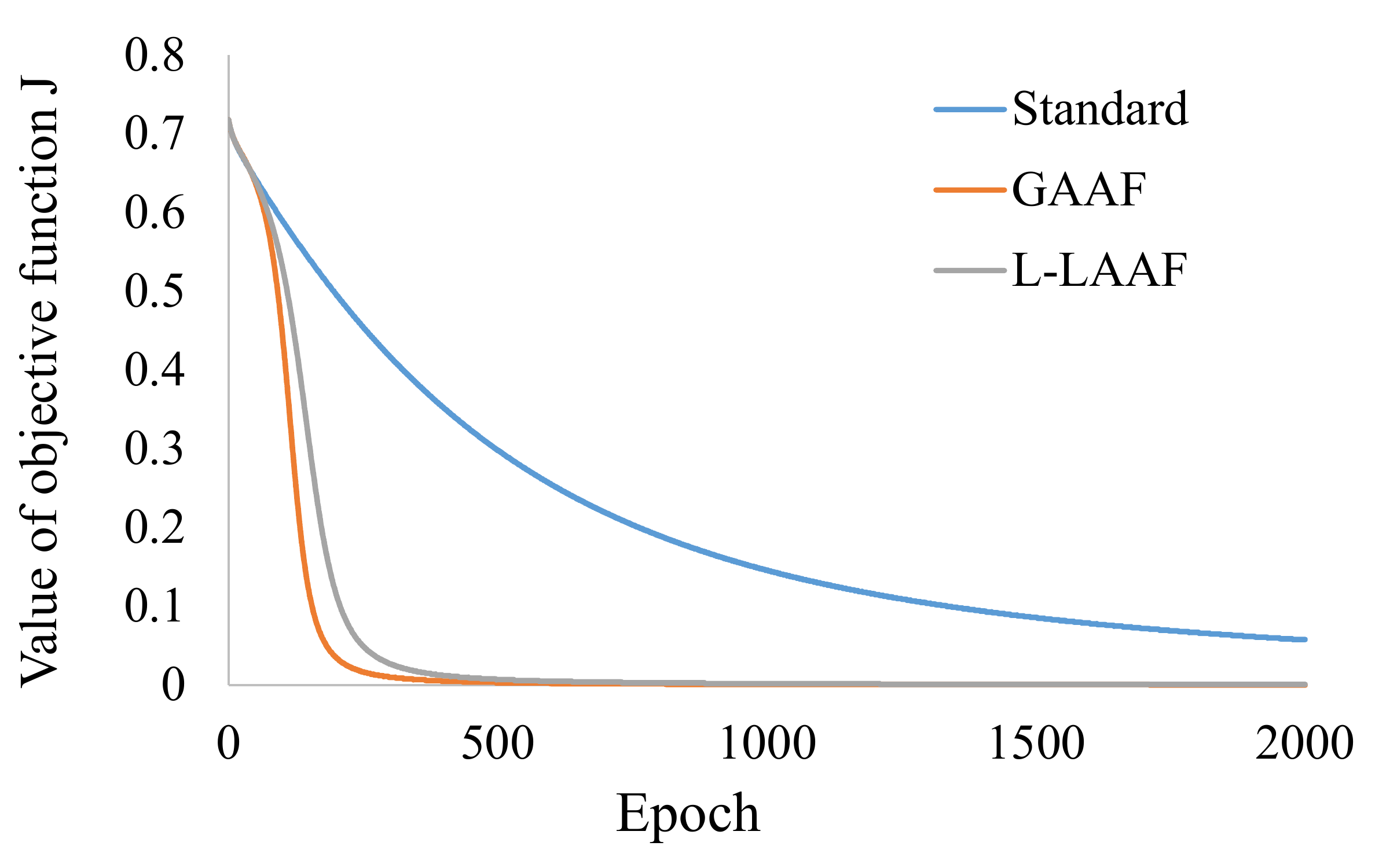}
  \includegraphics[width=\columnwidth,height=0.6\columnwidth]{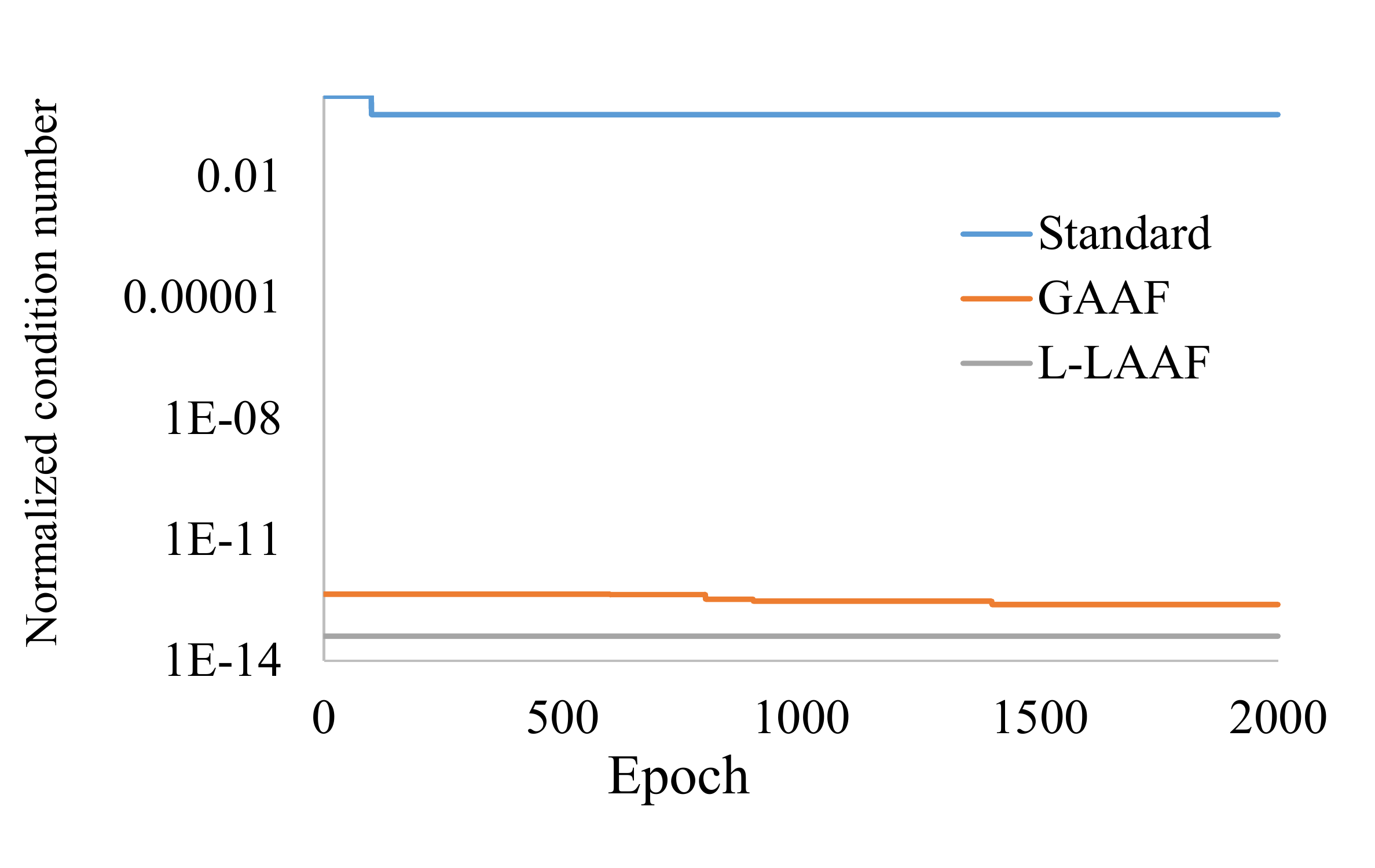}
  \vspace{-20pt}
  \caption{(ReLU, width=100)}
\end{subfigure}
\begin{subfigure}[b]{0.32\columnwidth}
  \includegraphics[width=\columnwidth,height=0.6\columnwidth]{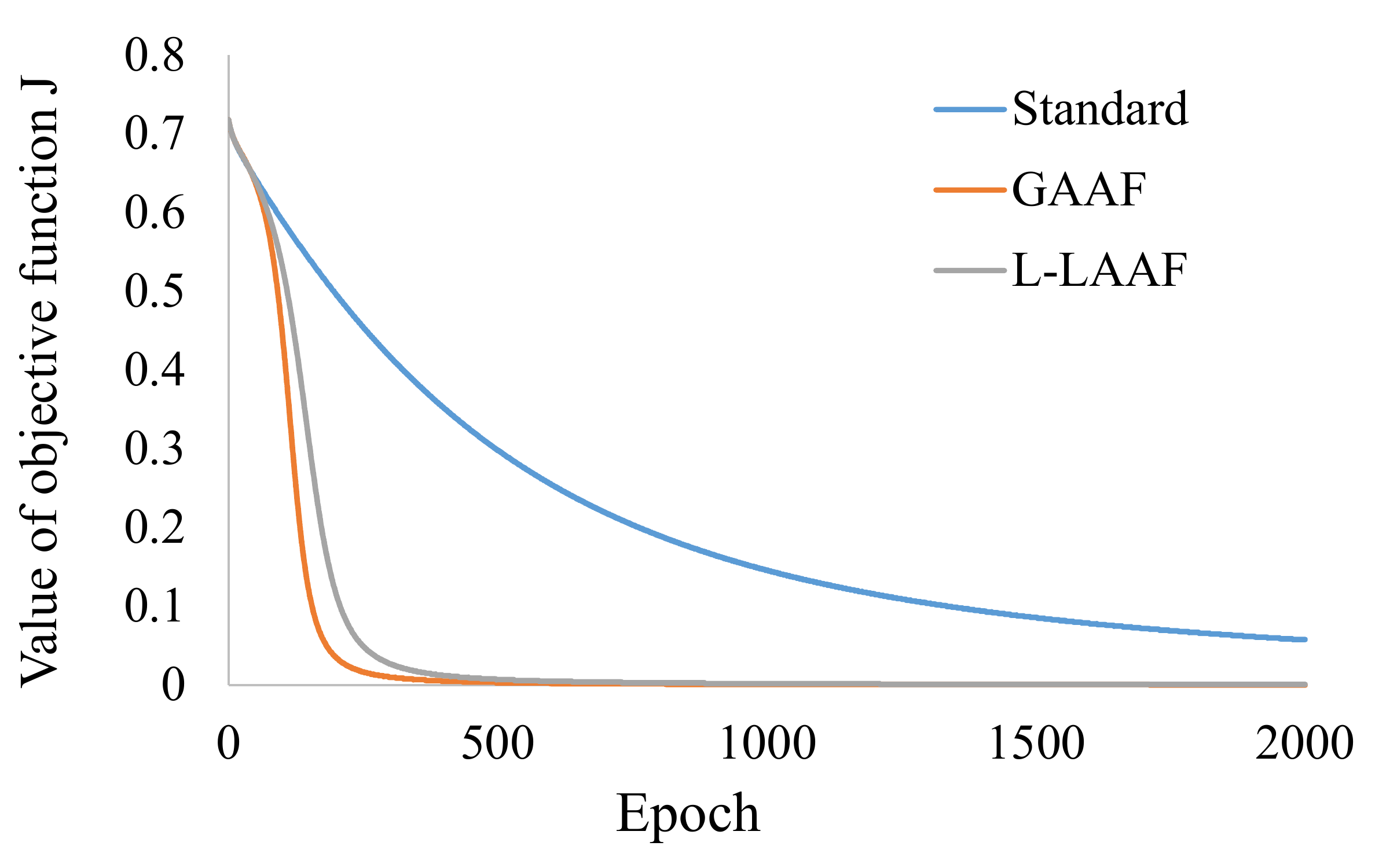}
  \includegraphics[width=\columnwidth,height=0.6\columnwidth]{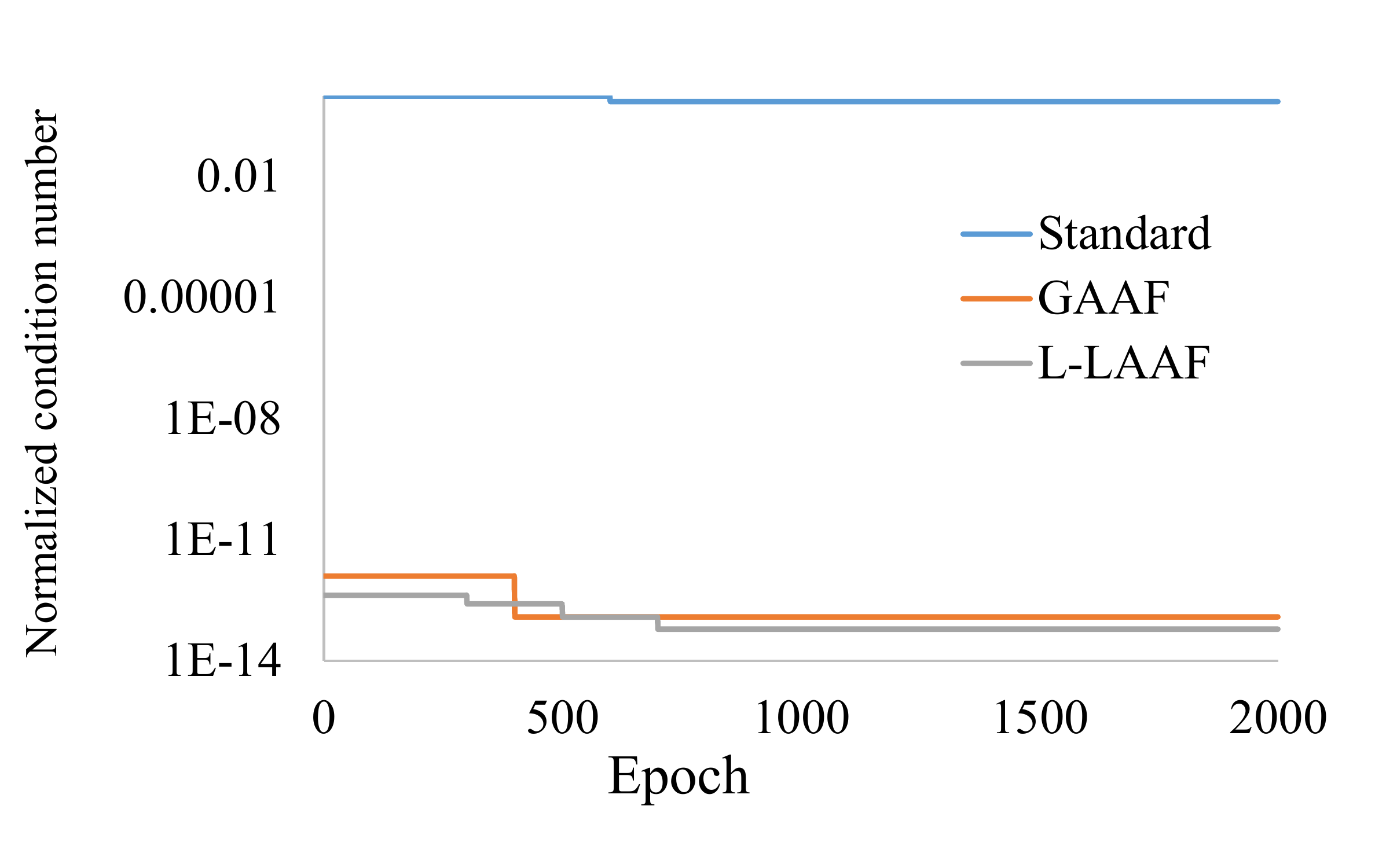}
  \vspace{-20pt}
  \caption{ (Softplus, width=100)} 
\end{subfigure}
\caption{Experiment under consideration: Effect of adaptive activations on objective value and  normalized condition number (i.e., the condition number divided by the initial condition number of the standard method). In each subfigure, width =  the number of units in  the hidden layer.}
\label{fig:condition}
\end{figure}

Figure \ref{fig:condition} illustrates the effect  of adaptive activation methods on the objective value and  the normalized condition number. It shows that the adaptive activation methods accelerated the convergence of the objective value while decreasing the condition number. The improvement of the condition number roughly coincided with the improvement of the objective value. The condition number of interest is (the largest singular value of $M$) / (the smallest singular value of $M$) where $M= G (\hat\bTheta)^{1/2} \nabla^2 J(\tilde\bTheta) G (\hat\bTheta)^{1/2}$ for adaptive activation function methods and $M=\nabla^2 J(\tilde\bTheta)$ for the standard method without adaptive activations. As this condition number gets smaller, the convergence rate of the objective value can improve when the matrix $M$ is normal \cite{bertsekas1999nonlinear}. The normalized condition numbers in each subfigure are the minimum condition numbers over the previous epochs, and were normalized by dividing the condition numbers by the initial condition numbers of the standard method. For this experiment, we set $n=1$, and we used a fixed dataset generated by sklearn.datasets.make\_circles with n\_samples = 1000, noise = 0.01, random\_state = 0, and factor = 0.7. This dataset is not linearly separable. We adopted the fully-connected neural network with a single hidden layer. The standard cross entropy loss was used for training and plots.

\section{Computational results}
In this section, we shall solve function approximation problem using deep NN and inverse PDE problems involving the two-dimensional Poisson's and inviscid Burgers equations using PINN algorithm with the proposed methods. Some standard deep learning benchmarks problems are also solved. The performance of the proposed methods is evaluated in terms of the convergence speed and the accuracy.

\subsection{Neural network approximation of nonlinear discontinuous function}
In this test case a deep neural network (without physics-informed part) is used to approximate a discontinuous function. Here, the loss function consists of the data mismatch and the slope recovery term.
\begin{figure} [htpb] 
\centering
\includegraphics[scale=0.48, angle = 0]{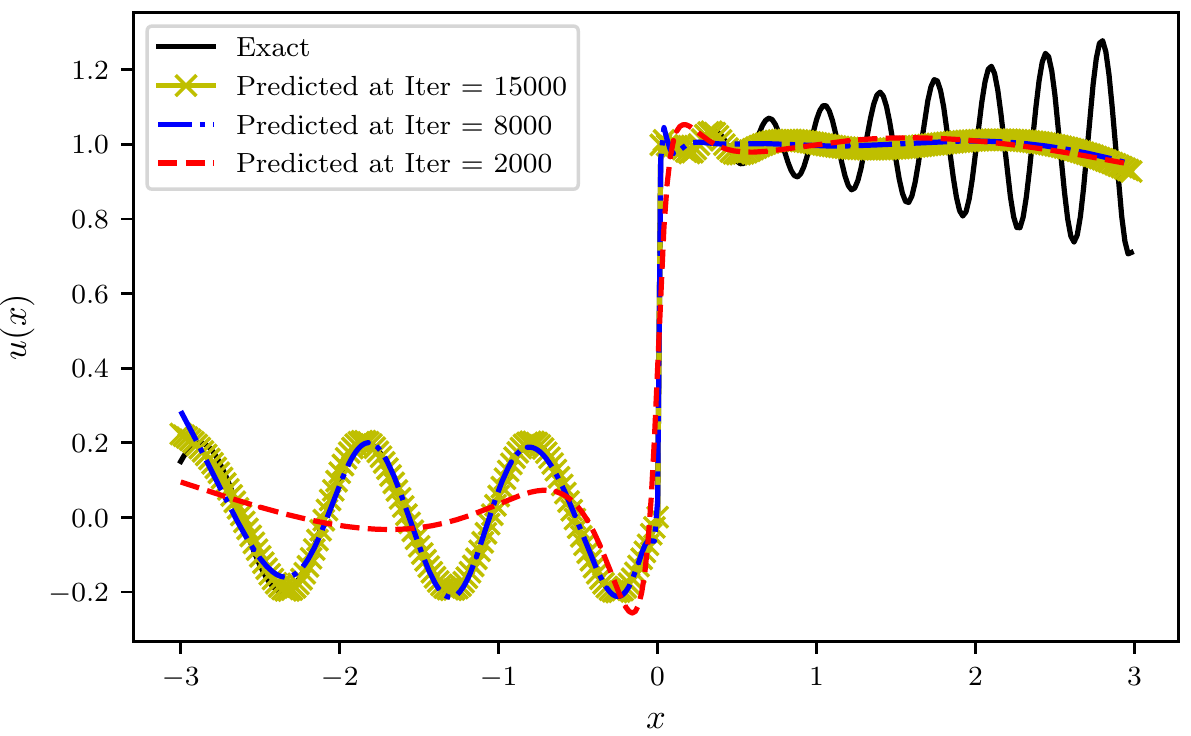}
\includegraphics[scale=0.48, angle = 0]{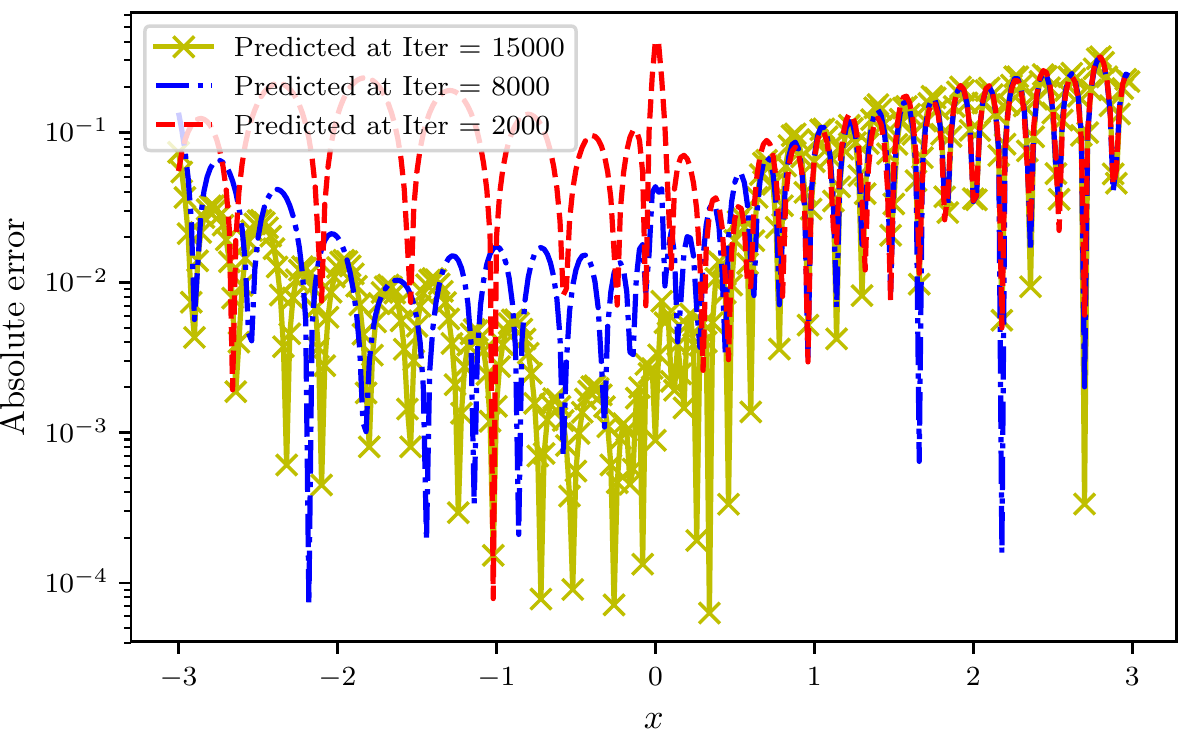}

\includegraphics[scale=0.48, angle = 0]{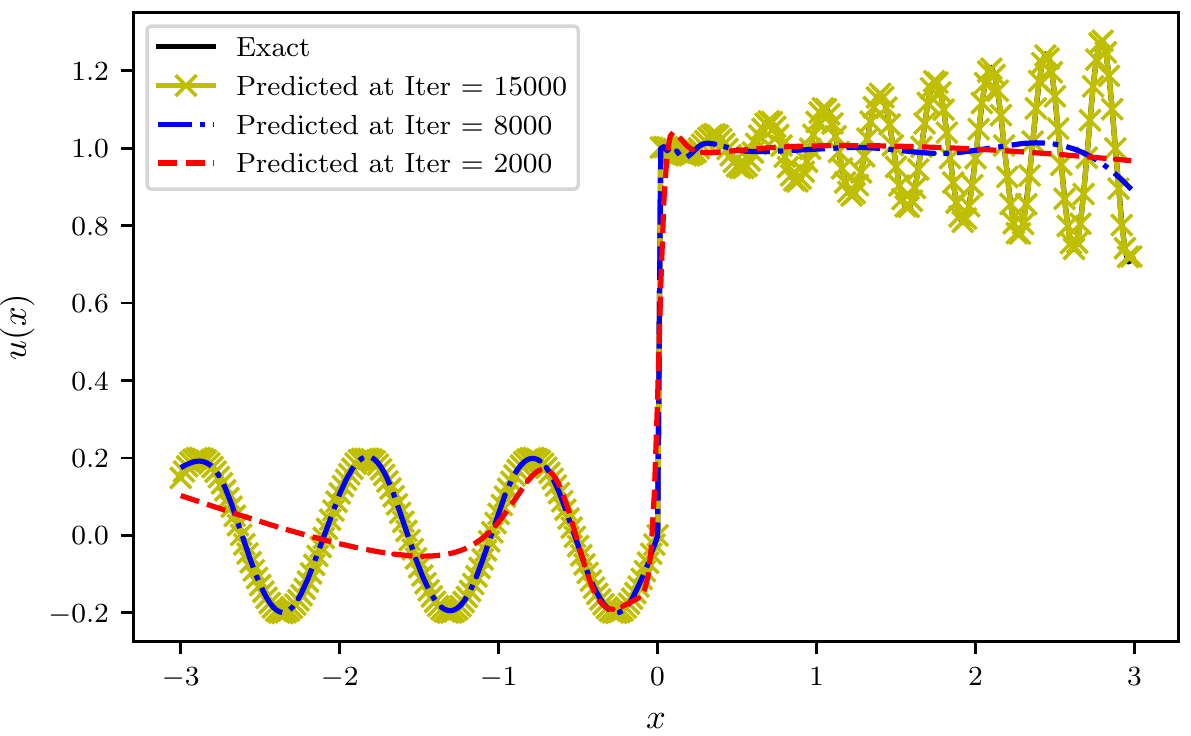}
\includegraphics[scale=0.48, angle = 0]{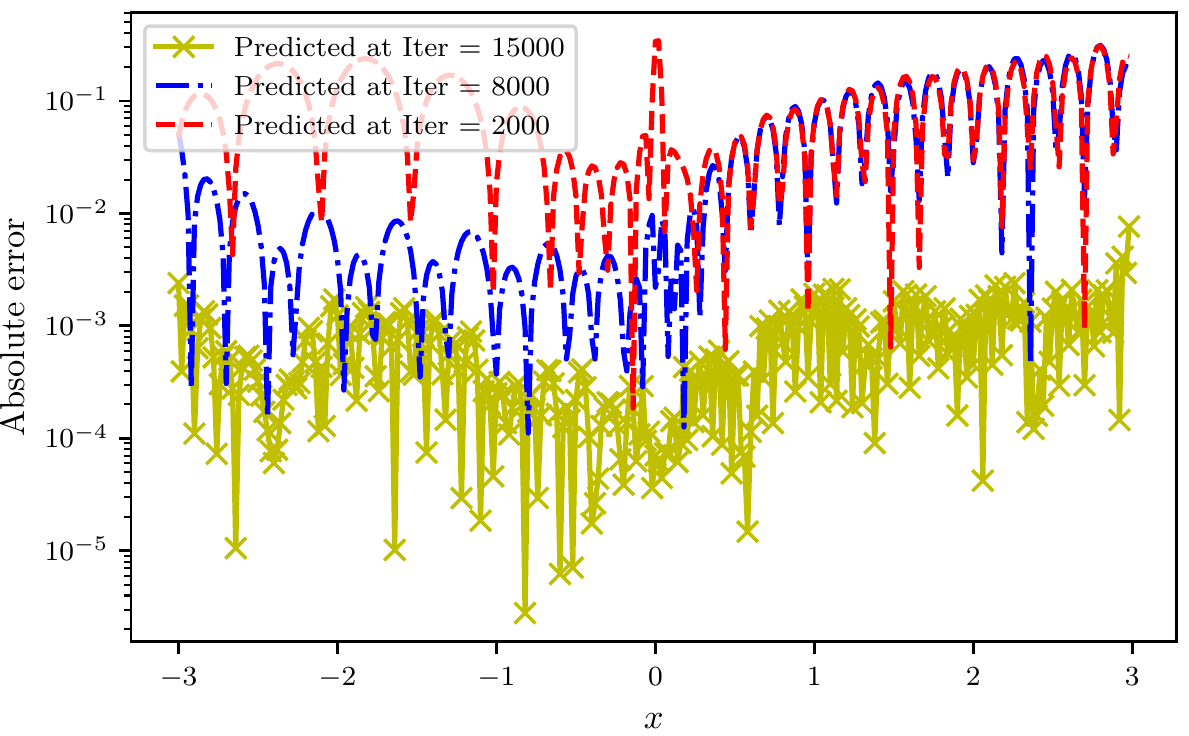}

\includegraphics[scale=0.48, angle = 0]{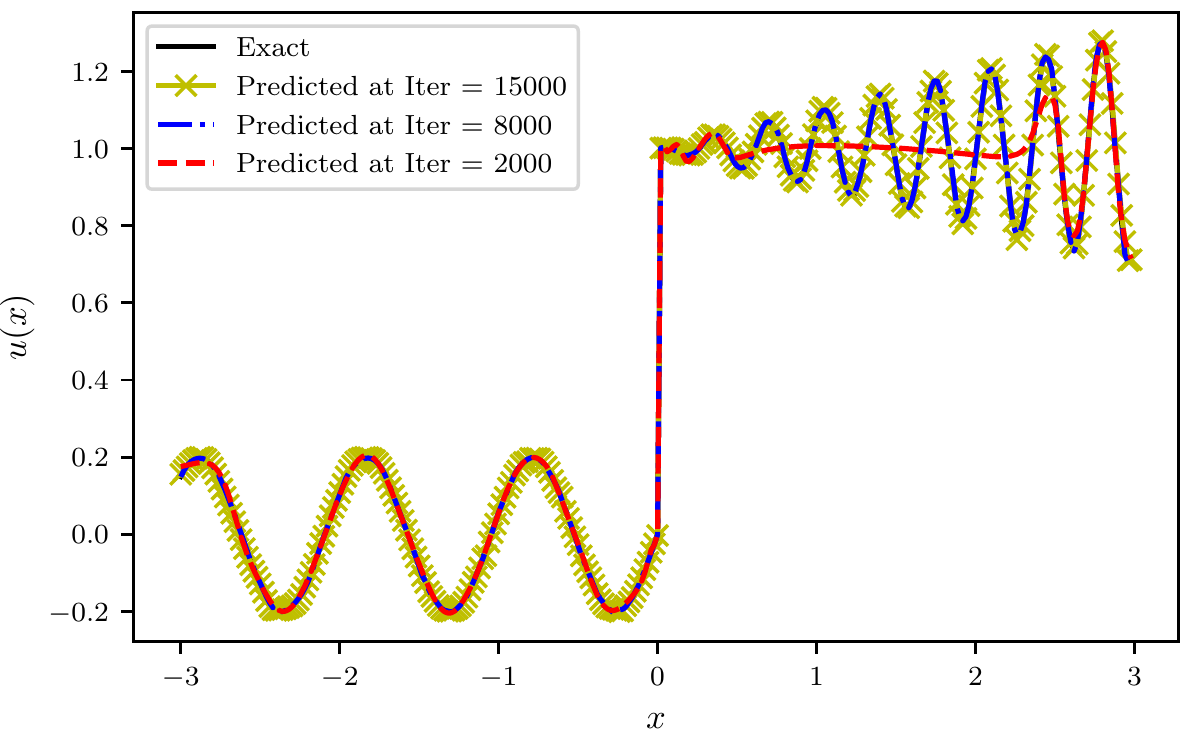}
\includegraphics[scale=0.48, angle = 0]{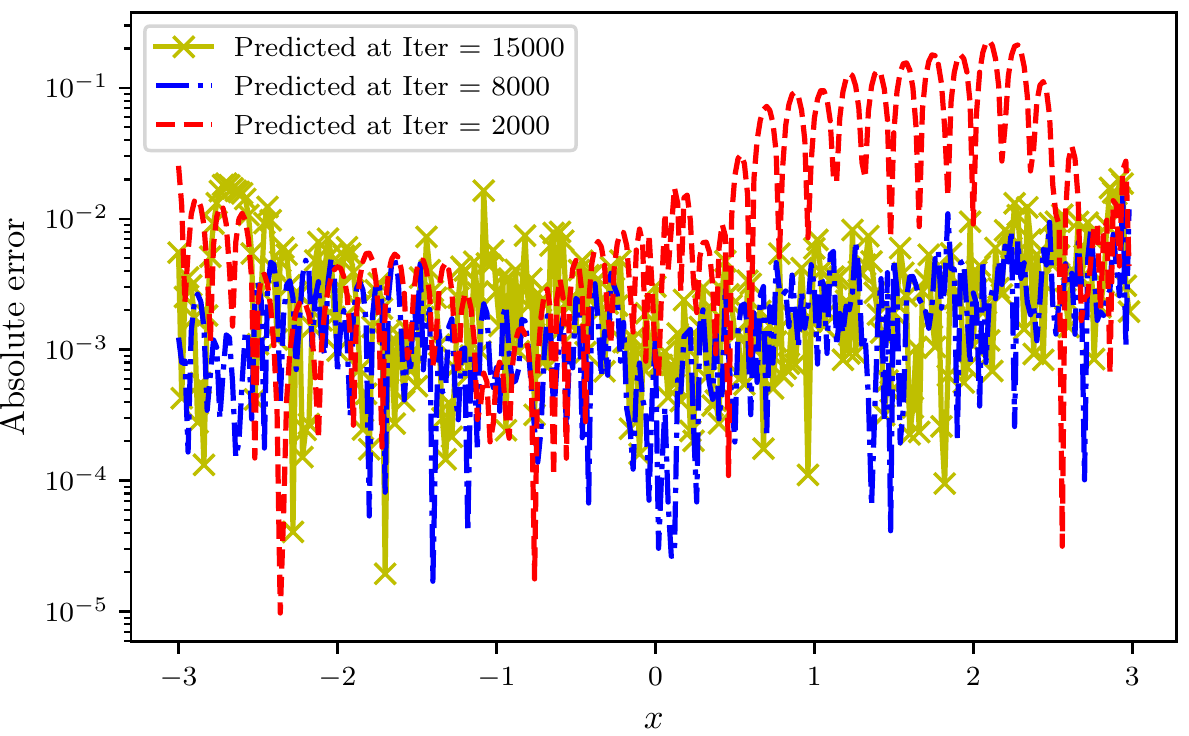}

\includegraphics[scale=0.48, angle = 0]{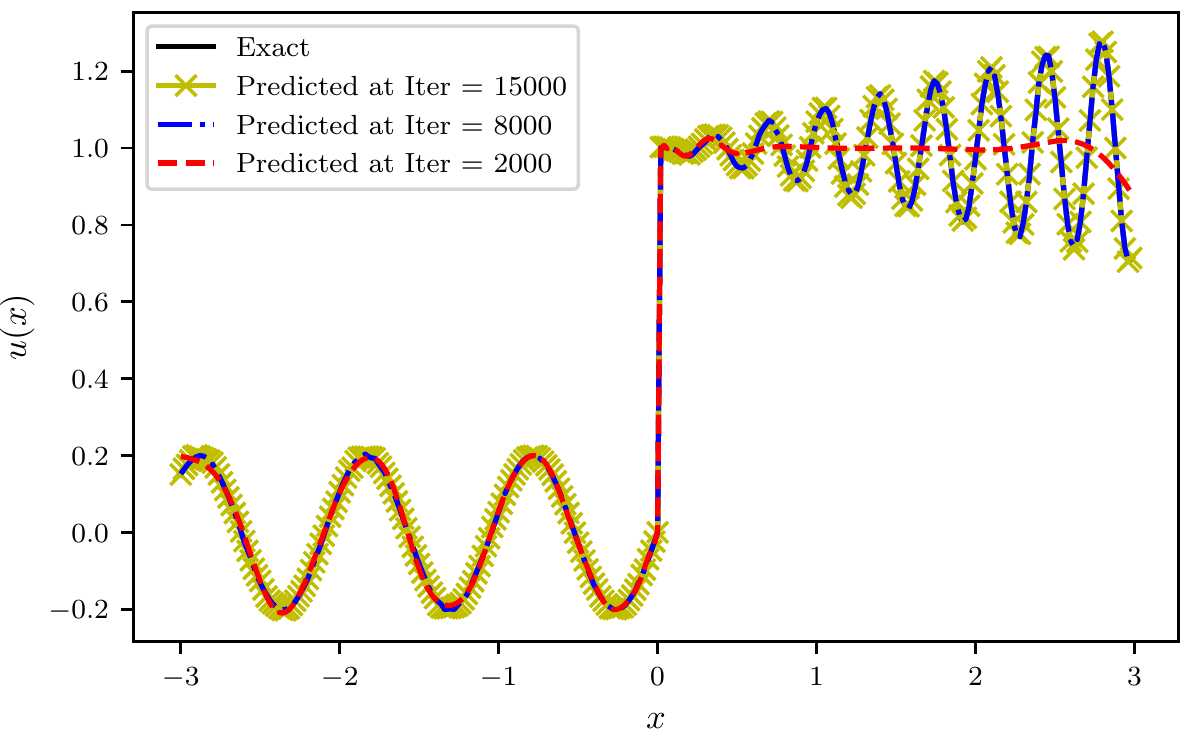}
\includegraphics[scale=0.48, angle = 0]{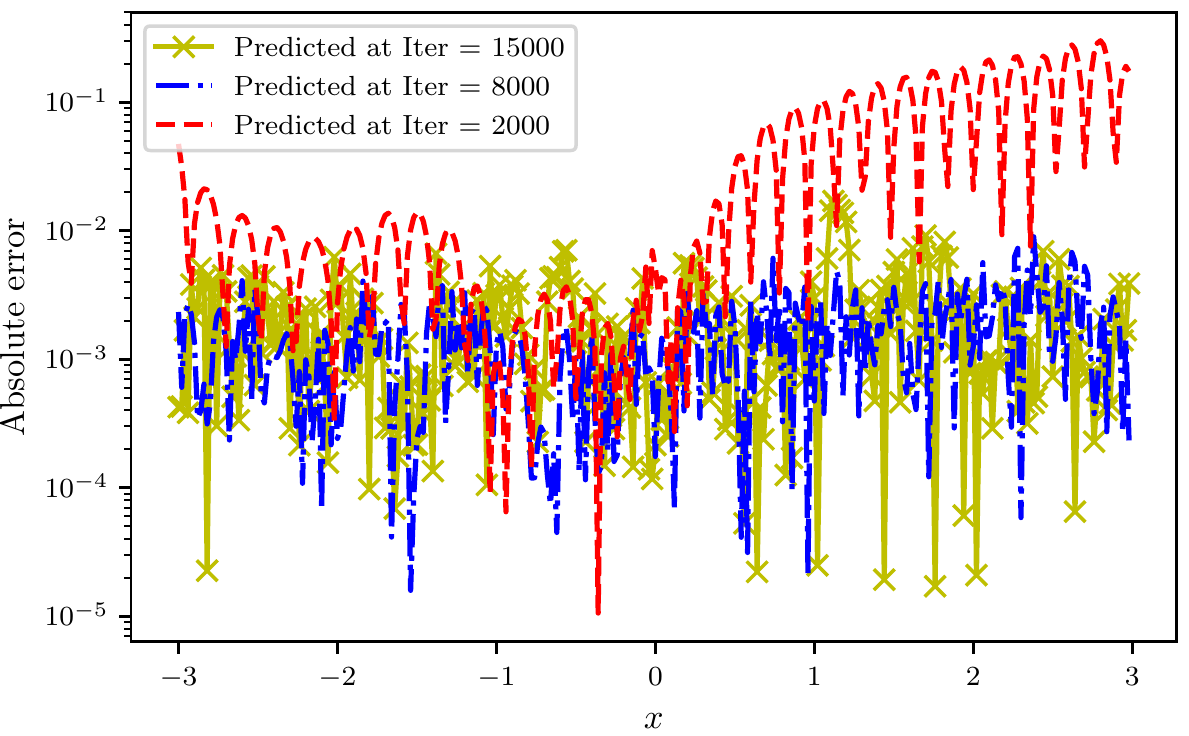}
\caption{Discontinuous function: Neural network approximation of function at 2000, 8000 and 15000 iterations using standard fixed activation (first row), GAAF (second row), L-LAAF (third row), and N-LAAF (fourth row) using the \textit{tanh} activation. The first column shows the solution. The second column gives the point-wise absolute error in the log scale for all the cases.}
\label{fig:Af}
\end{figure}
The following discontinuous function with discontinuity at $x = 0$ location is approximated by a deep neural network.
$$ u(x) = \begin{cases}
 0.2~\sin(6x)  & \text{If}~ x\leq 0, \\ 1+ 0.1~x\cos(18x)  & \text{Otherwise.} 
\end{cases}
 $$
This function contains both high and low frequency components along with the discontinuity. The domain is $[-3,~3]$ and the number of training points used is 300, which are chosen randomly. The activation function is \textit{tanh}, learning rate is 2.0e-4 and the number of hidden layers is four with 50 neurons in each layer. The scaling factor is 10 and both $W_u , W_a$ are unity. Figure \ref{fig:Af} shows the solution (first column) and point-wise absolute error in log scale (second column). The solution by the standard fixed activation function is given in the first row, the GAAF solution is given in second row, whereas the third and fourth row shows the solution given by L-LAAF and N-LAAF, respectively. We see that both L-LAAF and N-LAAF with slope recovery term accelerate training speed compared to other methods. We also note that, GAAF solution with the slope recovery term (not shown) is also comparable with the proposed methods in terms of training speed, and it can be considered as a new contribution due to involvement of the slope recovery term. 

\subsection{Inverse Problem: 2D Poisson's Equation}
In this example we shall identify the unknown parameter in the diffusion coefficient.
This example is taken from Pakravan et al \cite{Pak}, where variable diffusion coefficient parametrized as $\mathcal{D}(x; \alpha) = 1+ \alpha x$ needs to be evaluated, with randomly chosen values of $\alpha \in [0.05, 0.95]$. The computational domain is $\Omega \in [-1/\sqrt{2},~ 1/\sqrt{2}]^2$ and the governing Poisson's equation is given by
$$\nabla \cdot ([1+\alpha x] \nabla u) + x + y = 0, ~~~~~ (x,y) \in \Omega, $$
with boundary condition $u_b = \cos(\pi x)\cos(\pi y,)~(x,y) \in \partial \Omega$.

The feed forward neural network using three hidden layers with 30 neurons in each layer is trained over 500 generated solutions by randomly choosing the parameter $\alpha$ from the given range. The hyperbolic tangent activation function is used with 0.0008 learning rate. The value of scaling factor is unity in all the cases and $W_{\mathcal{F}} = 1, W_u = 10, W_a = 10$. Later we tested its performance on 50 independent solutions fields to identify the parameter $\alpha$ using the fixed activation, GAAF, L-LAAF and the N-LAAF without and with $2.5\%$ Gaussian noise. 
To show the convergence in the early training period, the results are plotted after 4000 iterations with the clean data and the data with noise. 
\begin{figure} [htpb] 
\centering 
\includegraphics[trim=2cm 1cm 2.2cm 0.5cm, clip=true, scale=0.57, angle = 0]{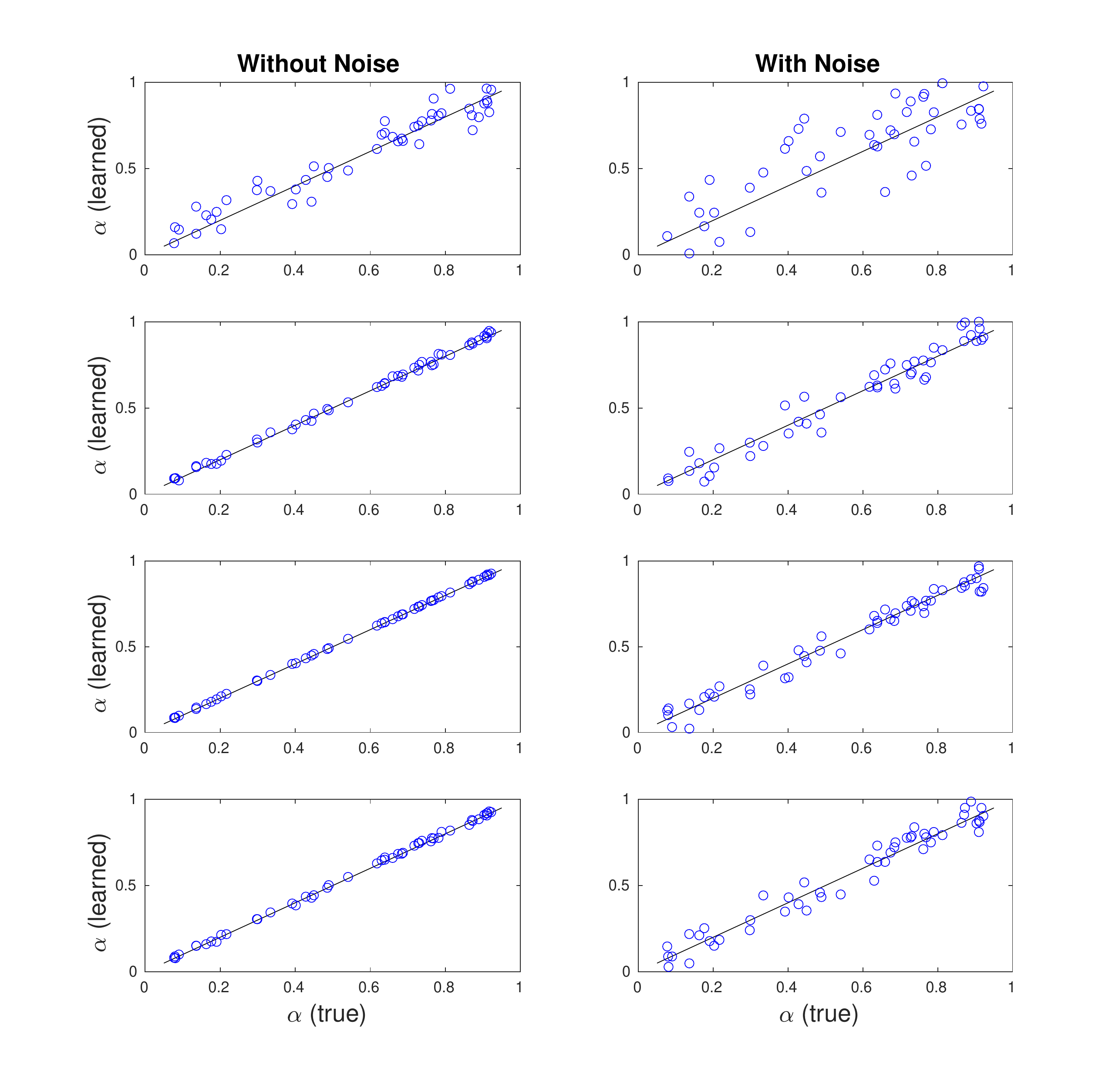}
\caption{Inverse problem 2D Poisson equation: The standard activation, GAAF, L-LAAF and the N-LAAF (top to bottom row) without noise (first column) and with Gaussian noise (second column).}
\label{fig:Poi2d}
\end{figure}
\begin{table}[htpb]
\begin{center}
\small \begin{tabular}{ccccc} \hline 
 &  Standard Acti. & GAAF &L-LAAF &N-LAAF 
 \\ \hline
Rel. $L_2$ error & 1.012e-1 &1.654e-2 & 7.328e-3 &1.482e-2 
 \\
 \hline 
 \end{tabular}
\caption{The relative $L_2$ error in all the cases without noise.}\label{Table2DPOis}
\end{center}
\end{table}
Figure \ref{fig:Poi2d} shows the results of the standard activation, GAAF, L-LAAF and the N-LAAF (top to bottom row) without noise (first column) and with Gaussian noise (second column). Both L-LAAF and N-LAAF perform better than GAAF where the learned parameters are in good agreement with the true values. Table \ref{Table2DPOis} shows the relative $L_2$ error in all the cases without noise, and among all methods, L-LAAF gives least error.

\subsection{Inverse Problem: 2D Inviscid Burgers Equation}
Our next example is the inverse problem of identification of viscosity coefficient in the Burgers equation. The solution of inviscid Burgers equation can be discontinuous even with sufficiently smooth initial condition. 
The two-dimensional inviscid Burgers equation is given as
$$u_t + uu_x + u_y = 0, ~~ x \in [-0.1,~1], y \in [0,~1],~ \text{and}~ t>0, $$
subject to boundary conditions $$u(x,0) = \begin{cases} a & \text{if}~ x<0, \\ b & \text{otherwise}, \end{cases}$$  $u(-0.2, y) = a$ and $u(1,y) = b, ~\forall y$.  
The exact solution for the case of $a = 2$ and $b = 0$ is given in \cite{ADJ_MRSU}, which has a steady oblique discontinuity at $x = 0$.

\begin{figure} [htpb] 
\centering 
\includegraphics[trim=0cm 0cm 0cm 0cm, clip=true, scale=0.36, angle = 0]{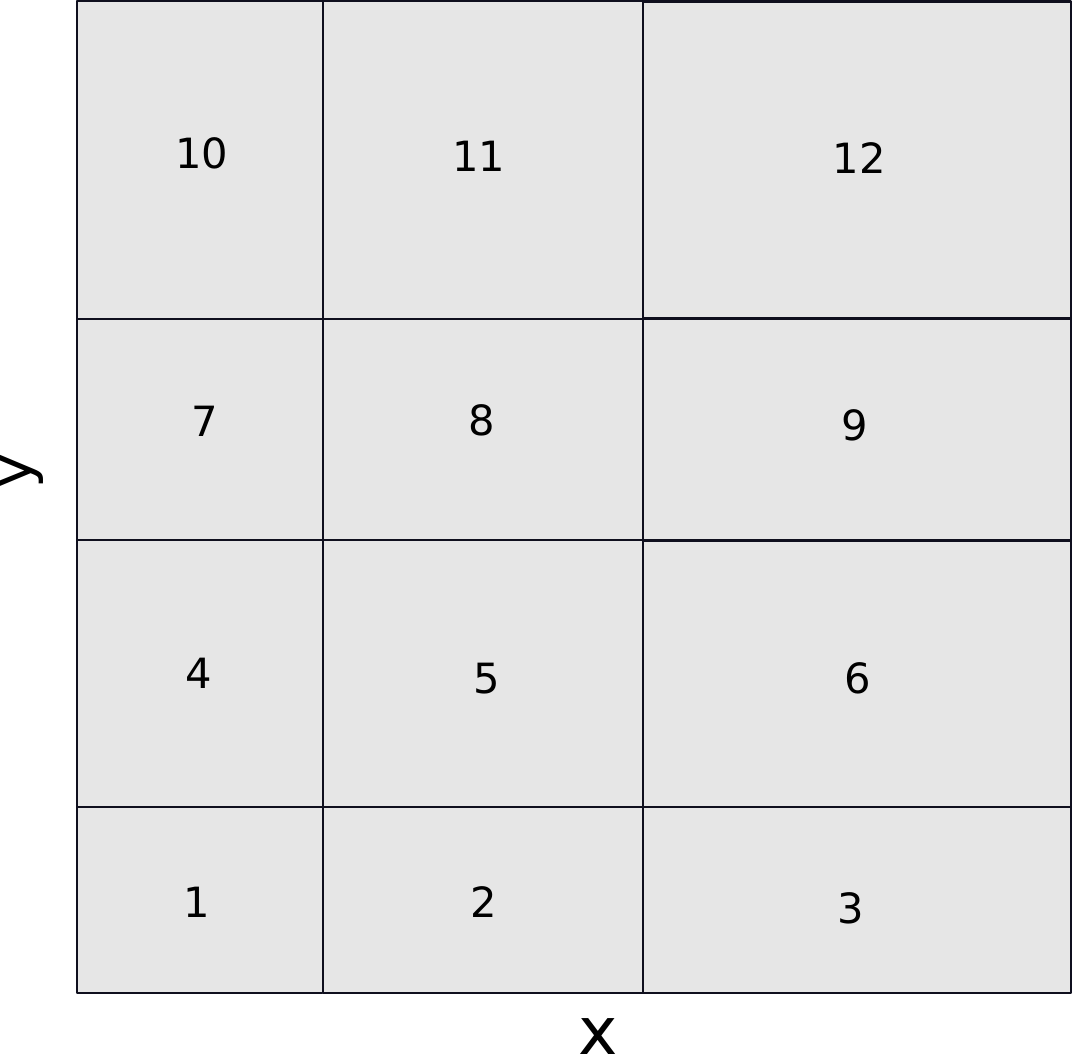}
\caption{Numbering of 12 sub-domains.}
\label{fig:cPINN12}
\end{figure}

In this work, we used the recently proposed conservative PINN (cPINN) method \cite{JK2}, which is basically a domain decomposition approach in PINN for conservation laws. The computational domain is divided into 12 sub-domains and a separate PINN is employed in each sub-domain which are working in tandem. The solutions in each sub-domains are stitched using the interface conditions, which include the enforcement of conservative flux and the average solution along the common interface. The division of computational domain into 12 sub-domains is shown in figure \ref{fig:cPINN12}. 
\begin{table}[htpb]
\begin{center}
\small \begin{tabular}{ccccccccccccc} \hline 
 Domain No. &  1 & 2&3 &4 & 5 & 6&7 &8 &  9 & 10&11 &12
 \\ \hline
$\#$ Layers & 6 &6 &3 &6 &6 &3& 6 &3 &3 &3 &3 &3
 \\
 $\#$ Neurons & 20 &20 &20 &20 &20 &20& 20 &20 &20 &20 &20 &20
 \\
$\#$ Residual Pts. & 2200 & 2200 & 400 & 600 & 2200 & 800 & 400 & 2200 & 2200 & 400 & 800 & 2200
 \\
 \hline 
 \end{tabular}
\caption{Neural network architecture in each sub-domains for the two-dimensional inviscid Burgers equation.}\label{Table2DinvBur}
\end{center}
\end{table}
The interface locations on $x$- and $y$-axes are [0.2, ~0.6] and  [0.25,~0.5,~0.75], respectively.

In cPINN algorithm, instead of supplying the original equation, the following parameterized viscous Burgers equation is supplied
$$u_t +  uu_x +  u_y = \nu ( u_{xx} + u_{yy}), $$
and we aim to identify the value of the viscosity coefficient $\nu$, which is zero for the inviscid Burgers equation.
\begin{figure} [htpb] 
\centering 
\includegraphics[trim=0.5cm 0cm 2cm 1.5cm, clip=true, scale=0.35, angle = 0]{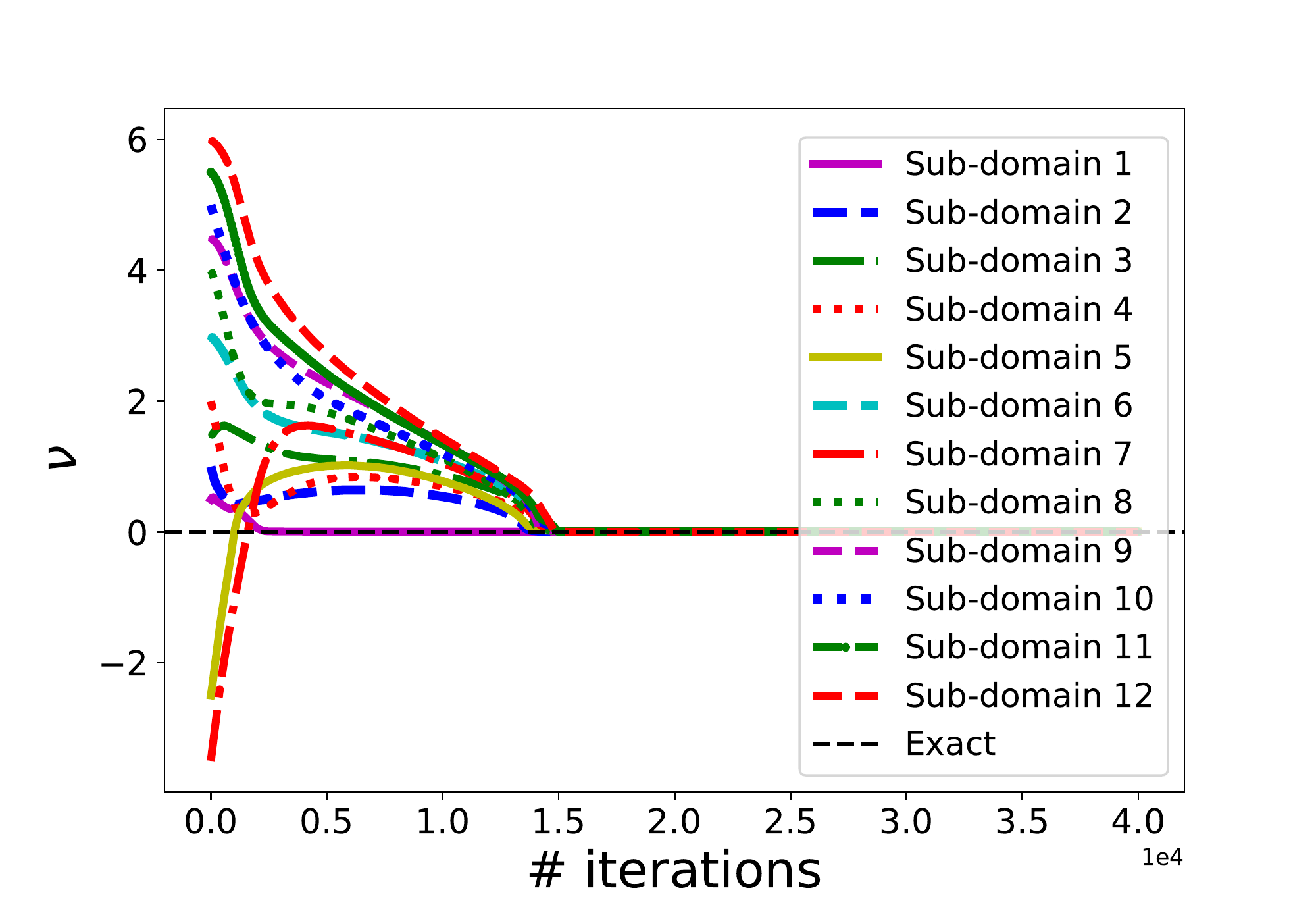}
\includegraphics[trim=0.5cm 2cm 2cm 2cm, clip=true, scale=0.3, angle = 0]{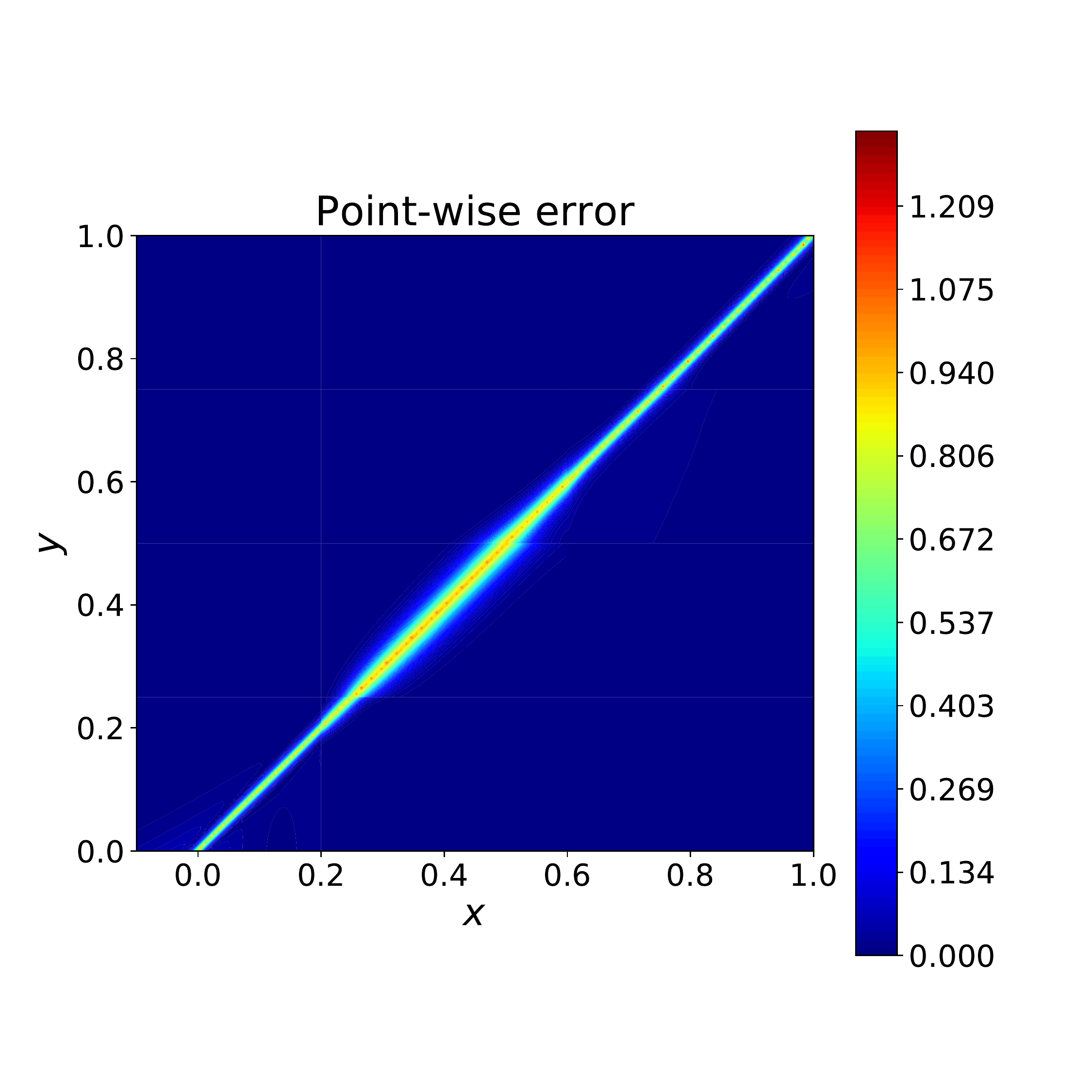}

\includegraphics[trim=0.5cm 0cm 2cm 1.5cm, clip=true, scale=0.35, angle = 0]{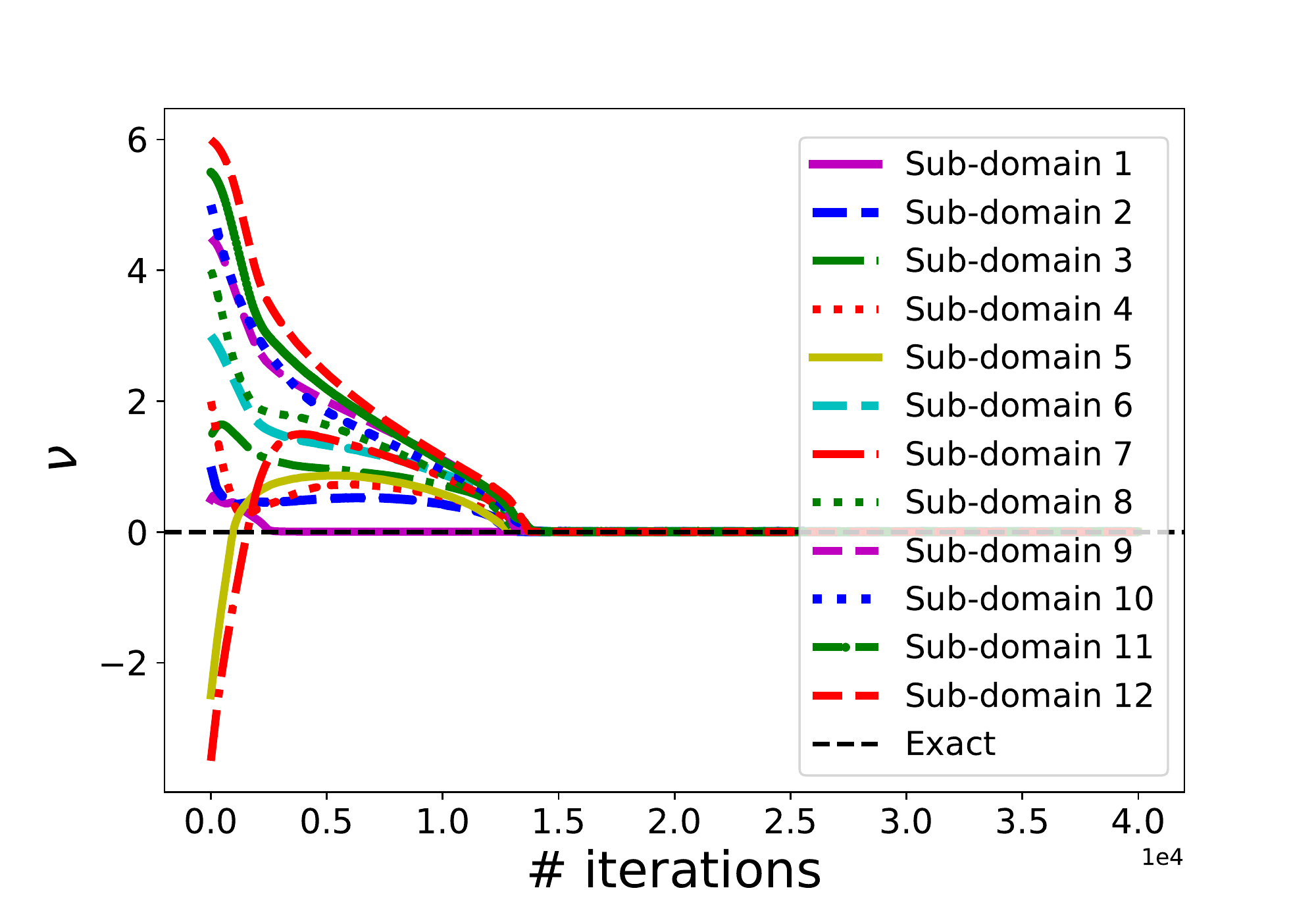}
\includegraphics[trim=0.5cm 2cm 2cm 2cm, clip=true, scale=0.3, angle = 0]{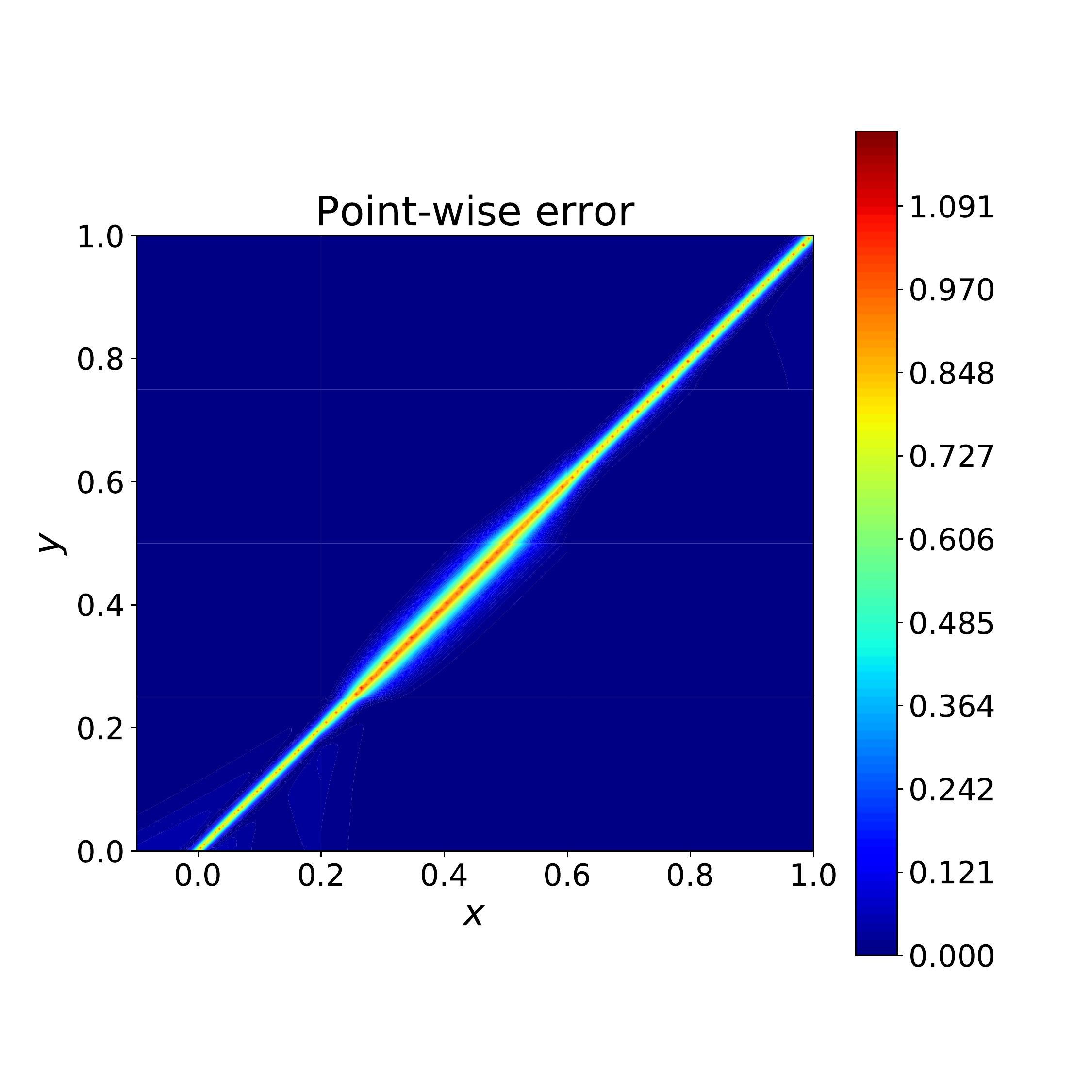}

\includegraphics[trim=0.5cm 0cm 2cm 1.5cm, clip=true, scale=0.35, angle = 0]{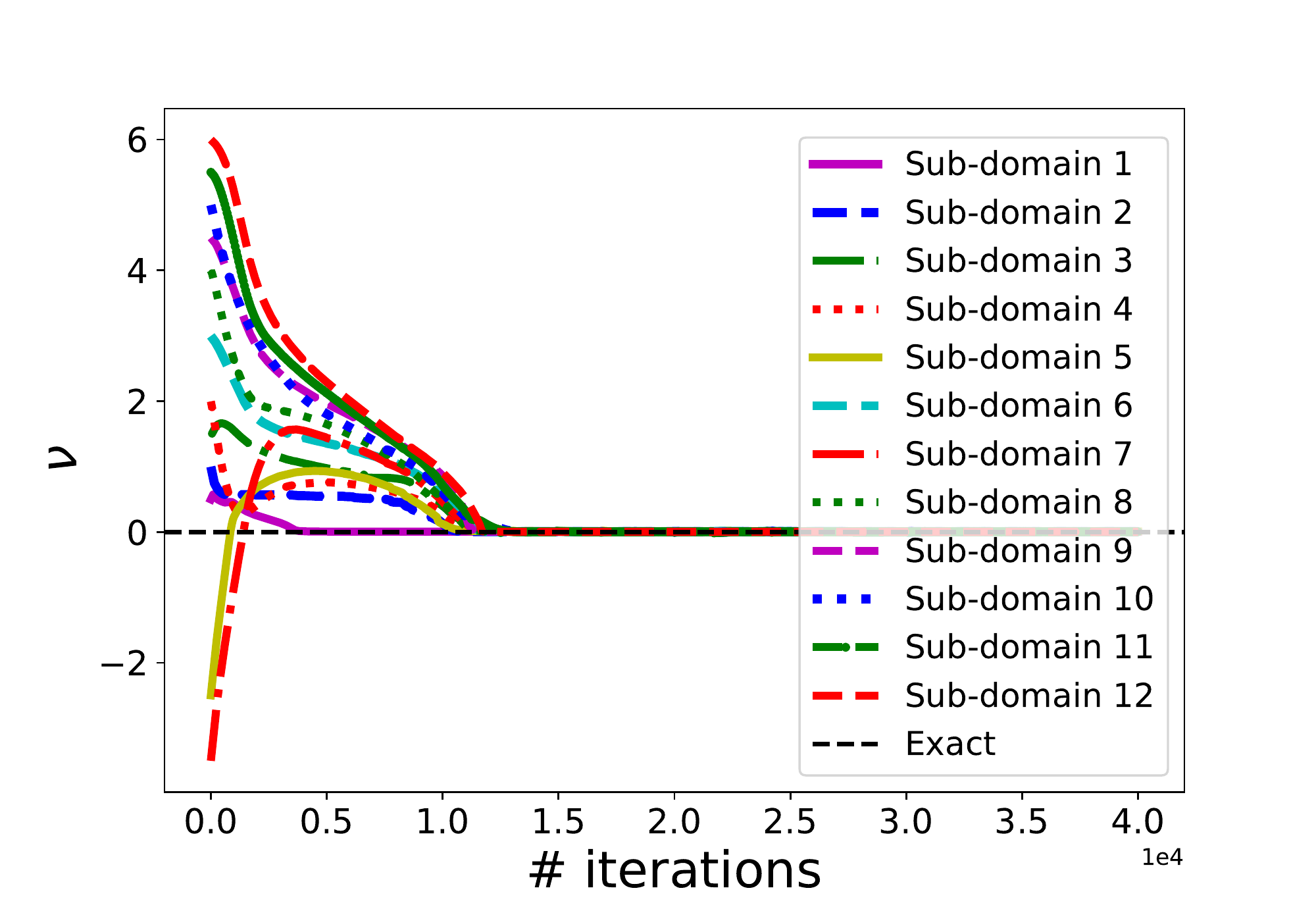}
\includegraphics[trim=0.5cm 2cm 2cm 2cm, clip=true, scale=0.3, angle = 0]{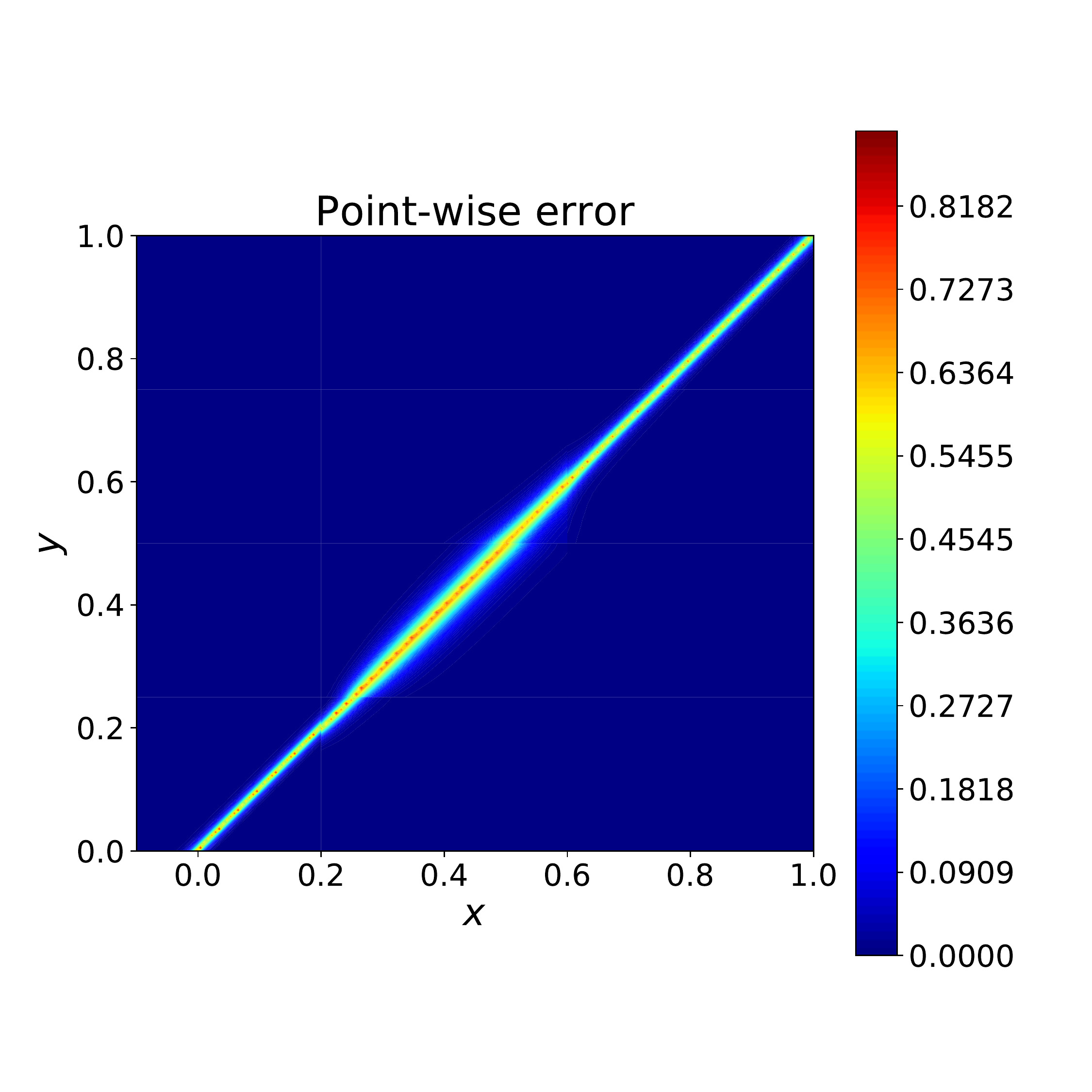}

\includegraphics[trim=0.5cm 0cm 2cm 1.5cm, clip=true, scale=0.35, angle = 0]{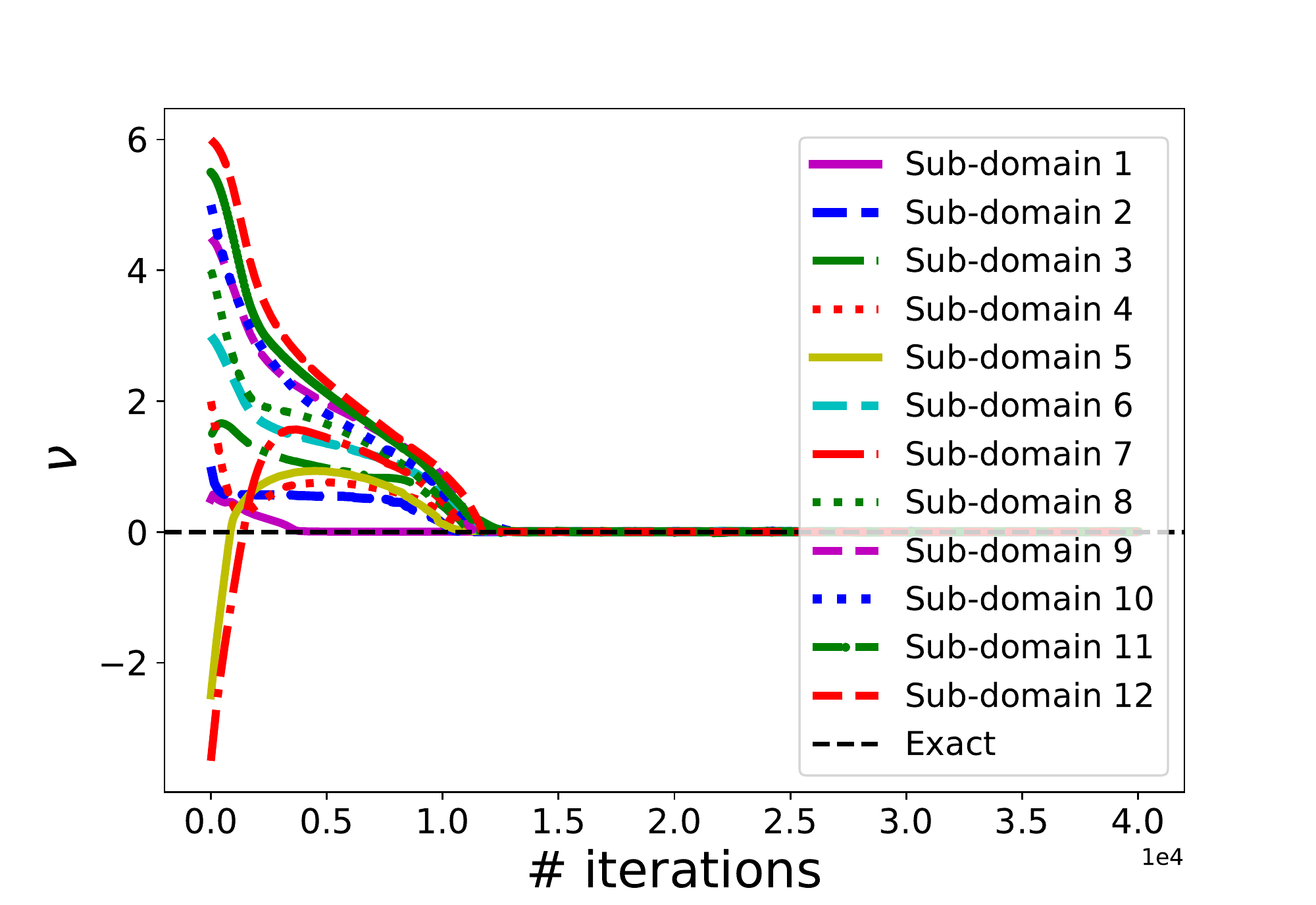}
\includegraphics[trim=0.5cm 2cm 2cm 2cm, clip=true, scale=0.3, angle = 0]{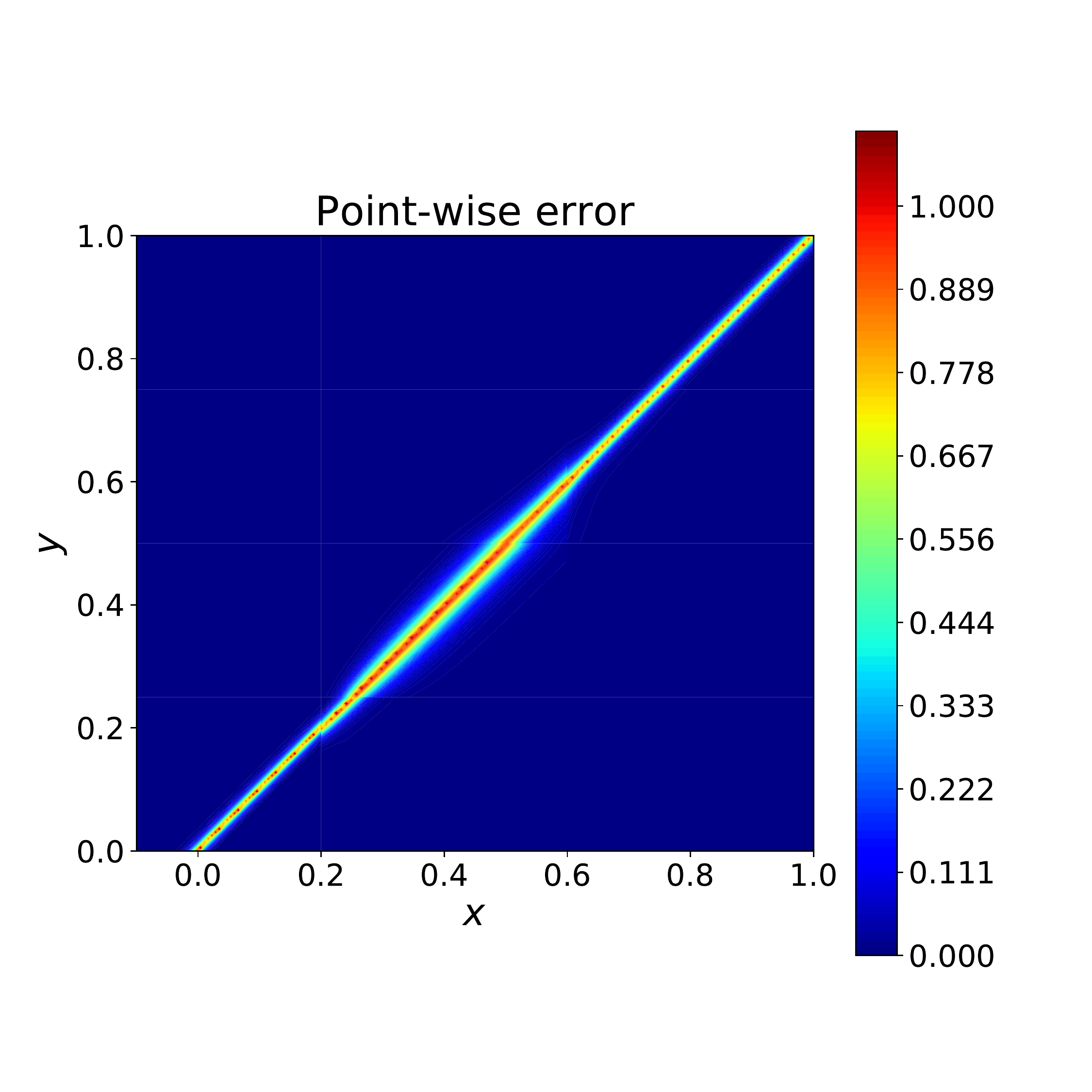}
\caption{Inverse problem 2D Burgers equation: Variation of $\nu$ (first column) and point-wise error (second column) for standard (fixed activation), GAAF, L-LAAF and N-LAAF (top to bottom row).}
\label{fig:cPINNInvBur}
\end{figure}
For this test case we used the hyperbolic tangent activation function, 0.0006 learning rate, scaling factor is 5 in all cases and table \ref{Table2DinvBur} gives the number of neurons, number of hidden layers and the number of residual points in each sub-domain. In this case, $W_{\mathcal{F}} = 1, W_u = 10, W_a = 20$ are used. The initial values of $\nu$ is chosen arbitrarily in all 12 sub-domains, which is [1, 2, 3, 4, -5, 6, -7, 8, 9, 10, 11, 12]/2, respectively. We note that cPINN is a robust method even in the case of negative values of viscosity that no standard solver would tolerate. 

All the simulations are performed upto 40k iterations. The number and locations of training data points (300 pts.) and residual points (8000 pts.) are fixed in all cases.
Figure \ref{fig:cPINNInvBur} shows the variation of $\nu$ (first column) and point-wise error (second column) for fixed activation, GAAF, L-LAAF and N-LAAF (top to bottom row) cases. In all cases, $\nu$ converges to its actual value, which is zero.
The cPINN algorithm using fixed activation function takes 14750 iterations for convergence whereas the GAAF, L-LAAF and the N-LAAF takes 13600, 11090 and 11170 iterations, respectively. Among all the cases, L-LAAF gives the smallest absolute point-wise error as shown in the figure. In this test case, both N-LAAF and L-LAAF with slope recovery term perform better than GAAF in terms of convergence speed and accuracy of the solution. 

\subsection{Standard Deep Learning Benchmark Problems}
The previous sub-section demonstrates the advantages of adaptive activation functions with PINN for physics related problems. One of the remaining questions is whether or not the advantage of adaptive activations remains with standard deep neural networks for other types of deep learning applications. To explore the question, this section presents numerical results with various standard benchmark problems in deep learning. 

\begin{figure}[hb!]
\begin{subfigure}[b]{0.02\columnwidth}
  \includegraphics[scale=0.17]{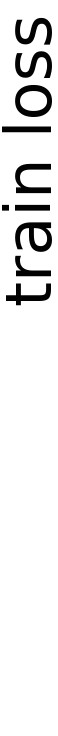}
\end{subfigure}
\begin{subfigure}[b]{0.45\columnwidth}
  \includegraphics[width=\columnwidth,height=0.55\columnwidth]{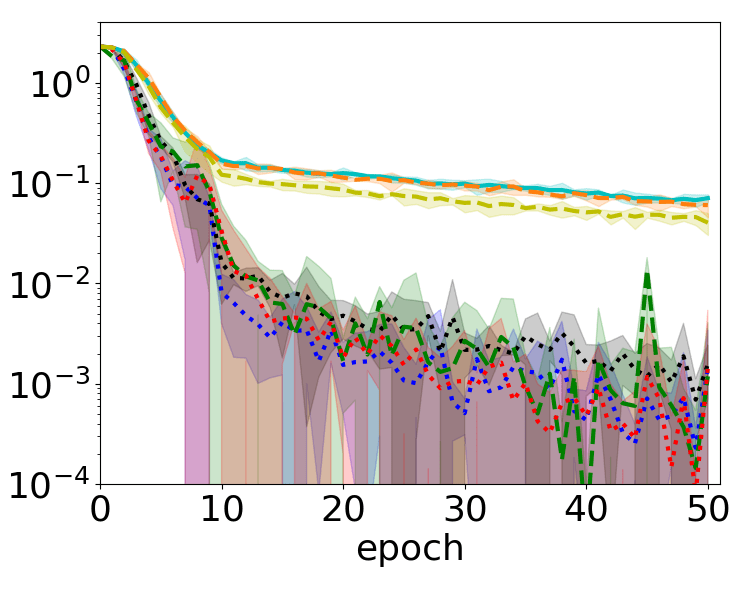}
  \vspace{-20pt}
  \caption{Semeion} 
\end{subfigure}
\begin{subfigure}[b]{0.45\columnwidth}
  \includegraphics[width=\columnwidth,height=0.55\columnwidth]{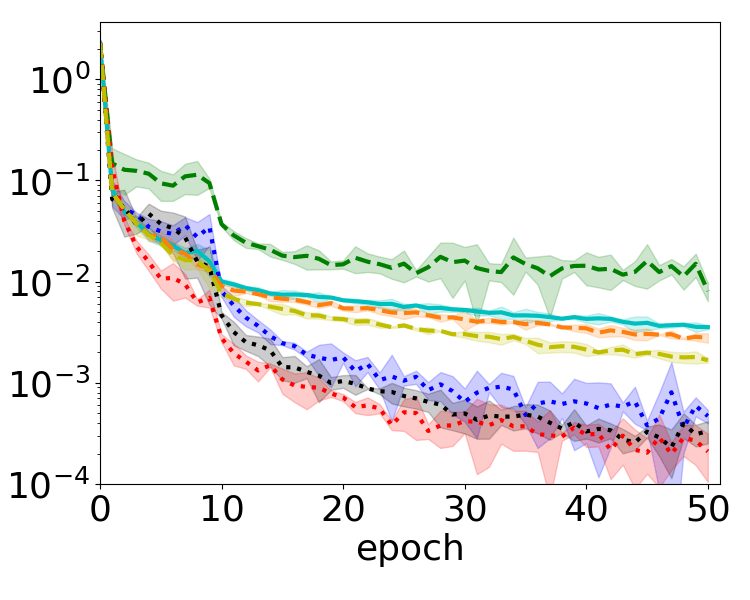}\llap{\shortstack{\includegraphics[scale=0.18]{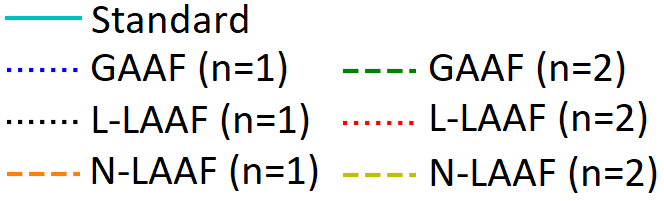}\\
         \rule{0ex}{1.00in}}\rule{0.25in}{0ex}}
  \vspace{-20pt}
  \caption{MNIST } 
\end{subfigure}

\begin{subfigure}[b]{0.02\columnwidth}
  \includegraphics[scale=0.17]{fig/labels/y_label}
\end{subfigure}
\begin{subfigure}[b]{0.45\columnwidth}
  \includegraphics[width=\columnwidth,height=0.55\columnwidth]{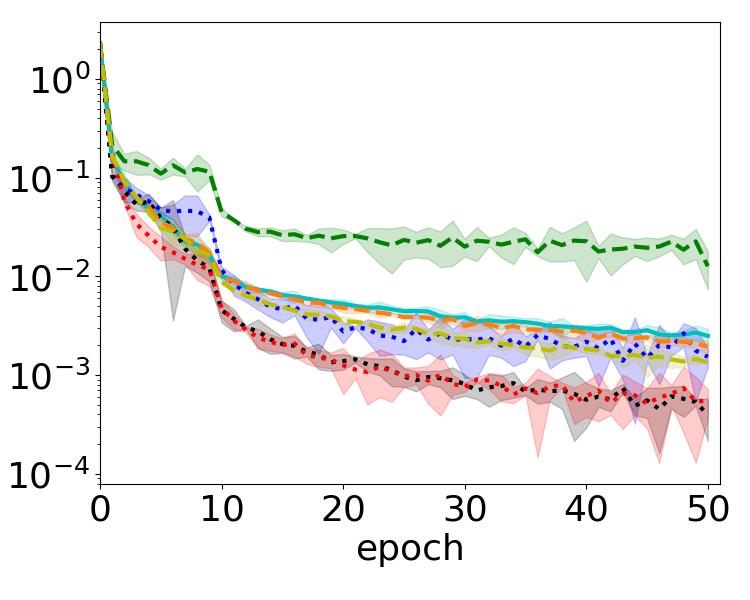}
  \vspace{-20pt}
  \caption{KMNIST} 
\end{subfigure}
\begin{subfigure}[b]{0.45\columnwidth}
  \includegraphics[width=\columnwidth,height=0.55\columnwidth]{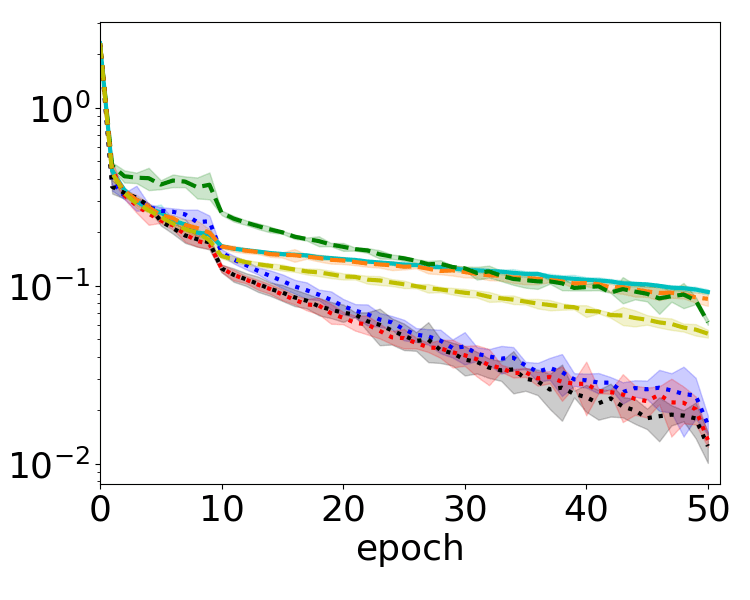}
  \vspace{-20pt}
  \caption{Fashion-MNIST} 
\end{subfigure}

\begin{subfigure}[b]{0.02\columnwidth}
  \includegraphics[scale=0.17]{fig/labels/y_label}
\end{subfigure}
\begin{subfigure}[b]{0.45\columnwidth}
  \includegraphics[width=\columnwidth,height=0.55\columnwidth]{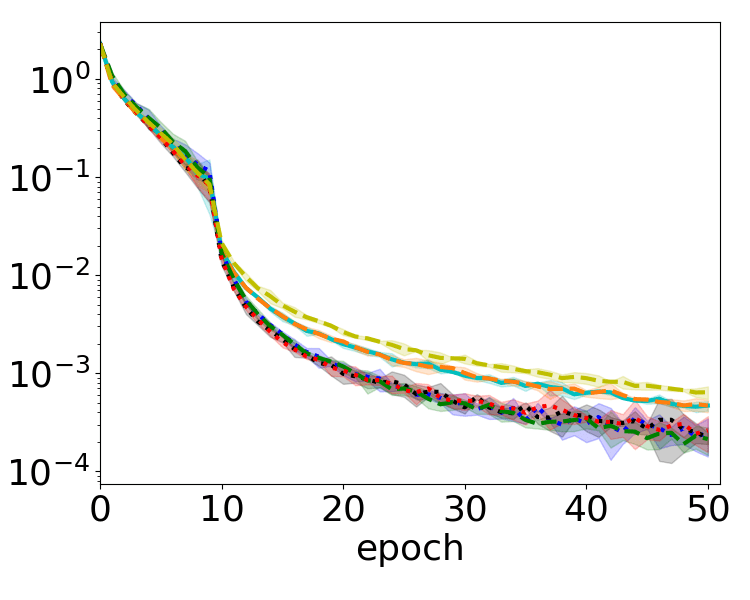}
  \vspace{-20pt}
  \caption{CIFAR-10} 
\end{subfigure}
\begin{subfigure}[b]{0.45\columnwidth}
  \includegraphics[width=\columnwidth,height=0.55\columnwidth]{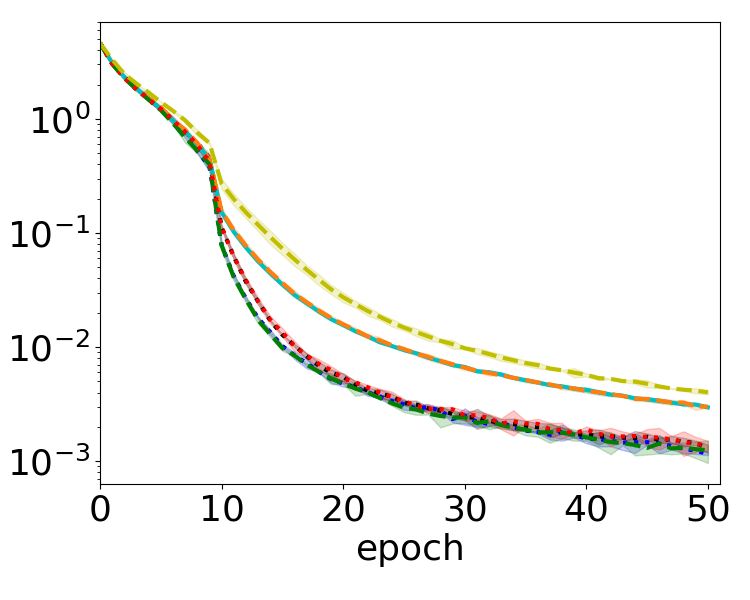}
  \vspace{-20pt}
  \caption{CIFAR-100} 
\end{subfigure}

\begin{subfigure}[b]{0.02\columnwidth}
  \includegraphics[scale=0.17]{fig/labels/y_label}
\end{subfigure}
\begin{subfigure}[b]{0.45\columnwidth}
  \includegraphics[width=\columnwidth,height=0.55\columnwidth]{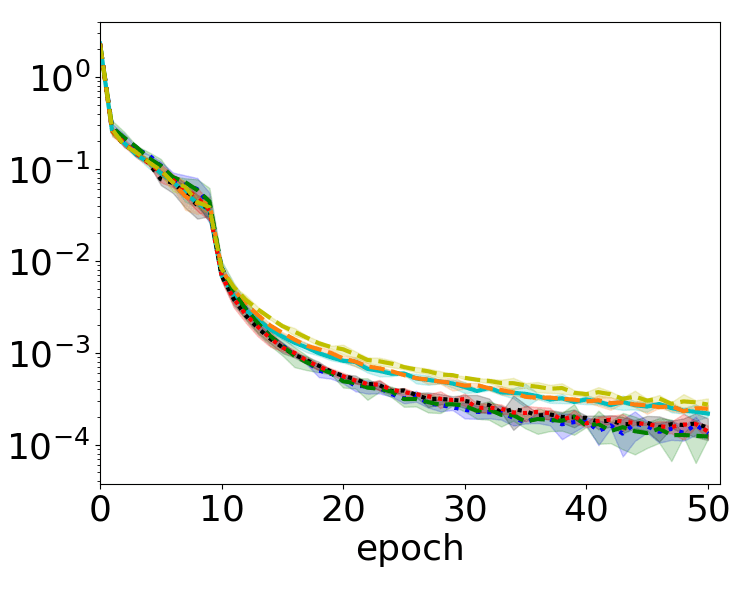}
  \vspace{-20pt}
  \caption{SVHN} 
\end{subfigure}
\caption{Training loss in log scale versus epoch without data augmentation. }
\label{fig:ml_plots_no_data_aug}
\end{figure}

\begin{figure}[ht!]
\begin{subfigure}[b]{0.02\columnwidth}
  \includegraphics[scale=0.17]{fig/labels/y_label}
\end{subfigure}
\begin{subfigure}[b]{0.45\columnwidth}
  \includegraphics[width=\columnwidth,height=0.55\columnwidth]{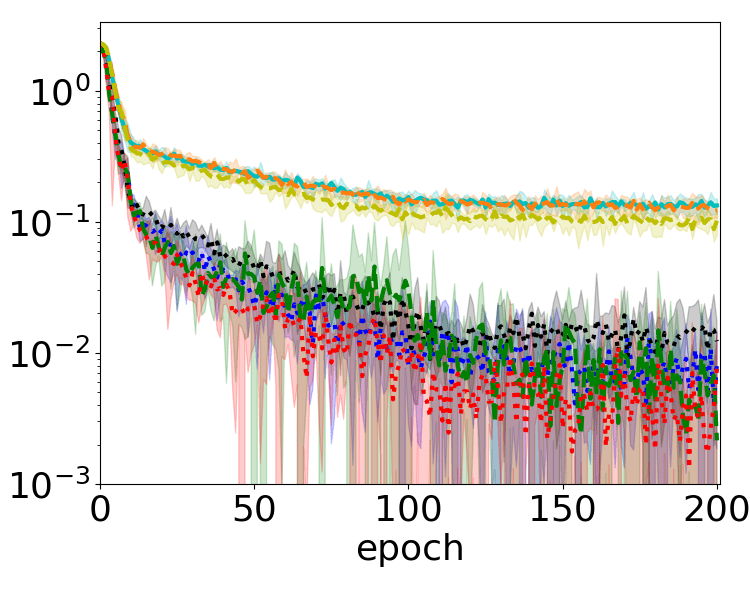}
  \vspace{-20pt}
  \caption{Semeion} 
\end{subfigure}
\begin{subfigure}[b]{0.45\columnwidth}
  \includegraphics[width=\columnwidth,height=0.6\columnwidth]{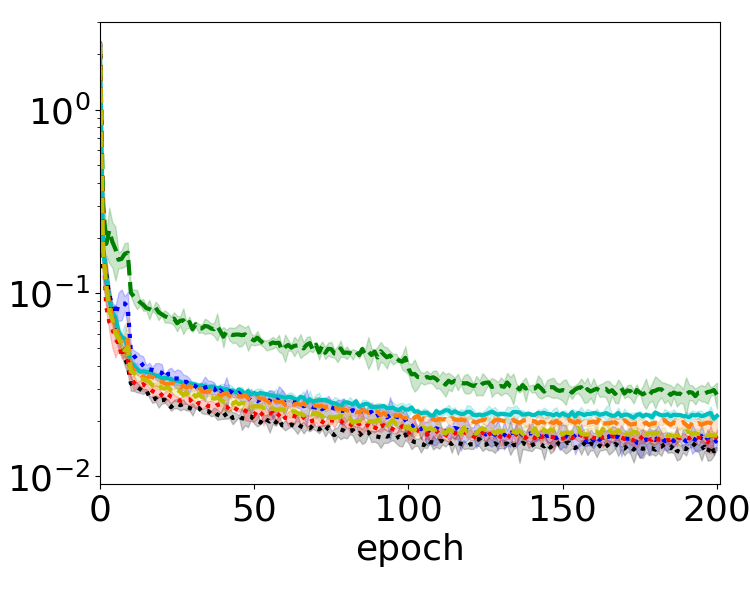}\llap{\shortstack{\includegraphics[scale=0.22]{fig/labels/legend}\\
        \rule{0ex}{1.00in}}\rule{0.25in}{0ex}}
  \vspace{-20pt}
  \caption{MNIST} 
\end{subfigure}

\begin{subfigure}[b]{0.02\columnwidth}
  \includegraphics[scale=0.17]{fig/labels/y_label}
\end{subfigure}  
\begin{subfigure}[b]{0.45\columnwidth}
  \includegraphics[width=\columnwidth,height=0.55\columnwidth]{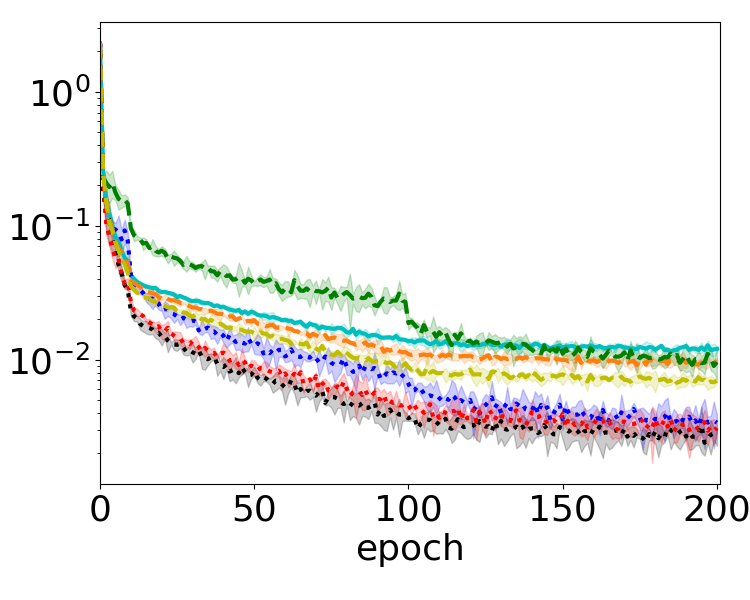}
  \vspace{-20pt}
  \caption{KMNIST} 
\end{subfigure}
\begin{subfigure}[b]{0.45\columnwidth}
  \includegraphics[width=\columnwidth,height=0.55\columnwidth]{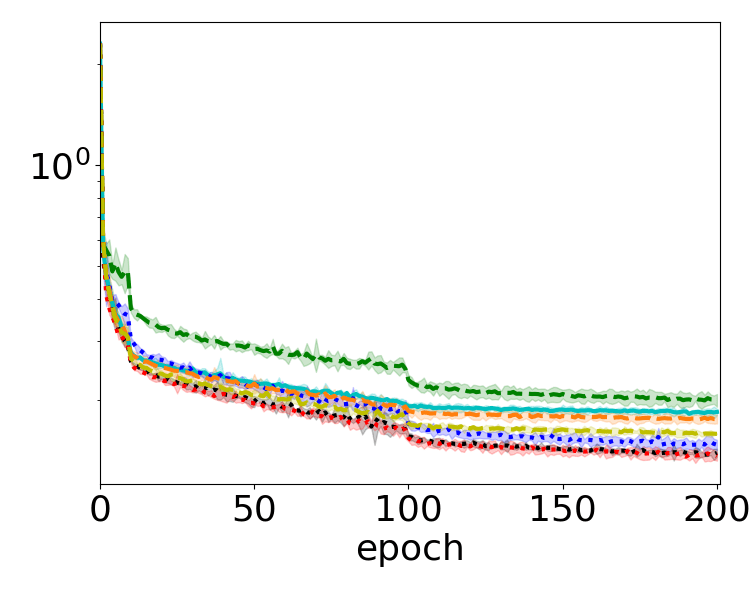}
  \vspace{-20pt}
  \caption{Fashion-MNIST} 
\end{subfigure}

\begin{subfigure}[b]{0.02\columnwidth}
  \includegraphics[scale=0.17]{fig/labels/y_label}
\end{subfigure}
\begin{subfigure}[b]{0.45\columnwidth}
  \includegraphics[width=\columnwidth,height=0.55\columnwidth]{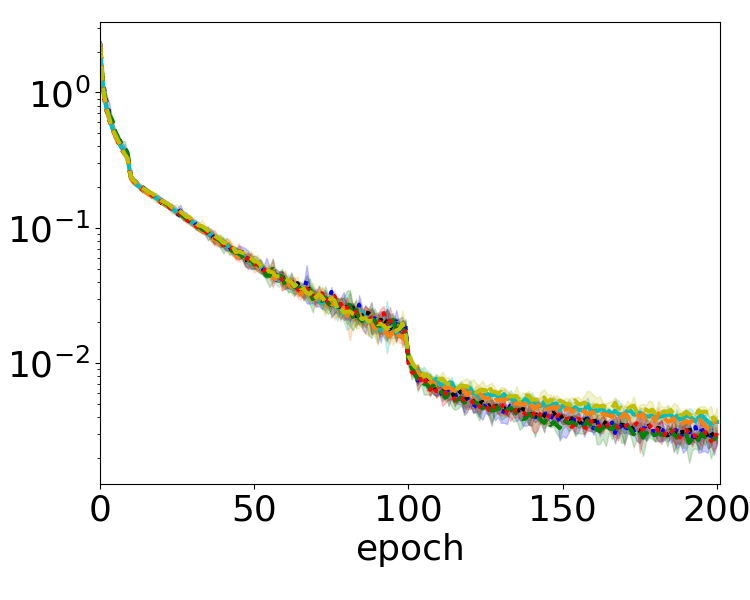}
  \vspace{-20pt}
  \caption{CIFAR-10} 
\end{subfigure}
\begin{subfigure}[b]{0.45\columnwidth}
  \includegraphics[width=\columnwidth,height=0.55\columnwidth]{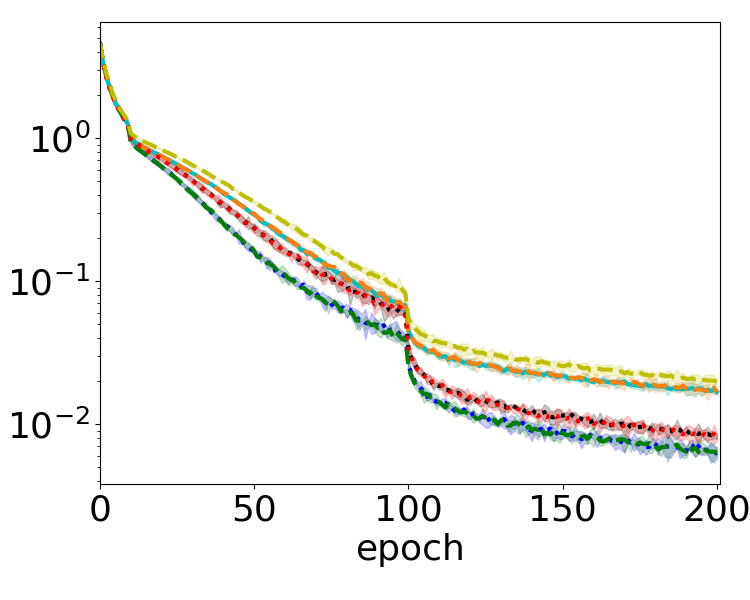}
  \vspace{-20pt}
  \caption{CIFAR-100} 
\end{subfigure}

\begin{subfigure}[b]{0.02\columnwidth}
  \includegraphics[scale=0.17]{fig/labels/y_label}
\end{subfigure}
\begin{subfigure}[b]{0.45\columnwidth}
  \includegraphics[width=\columnwidth,height=0.55\columnwidth]{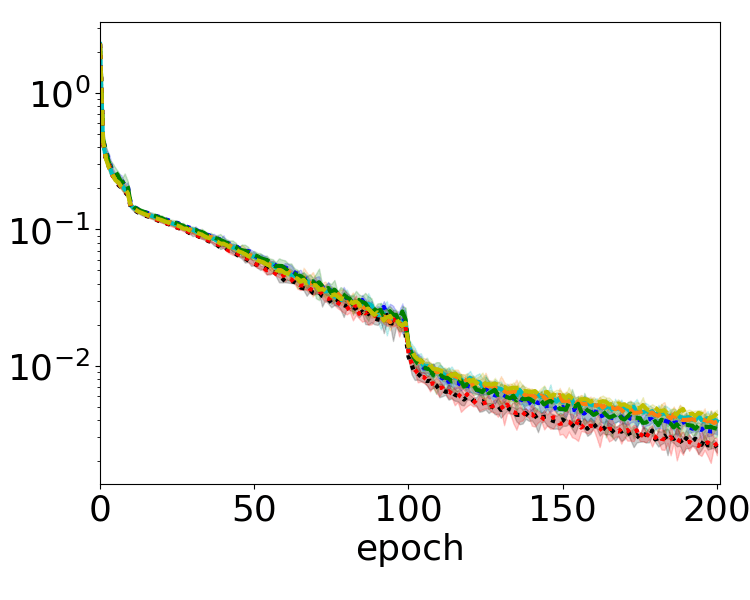}
  \vspace{-20pt}
  \caption{SVHN} 
\end{subfigure}
\caption{Training loss in log scale versus epoch with data augmentation. }
\label{fig:ml_plots_data_aug}
\end{figure}

MNIST \cite{LeCun}, Fashion-MNIST \cite{FMNIST} and KMNIST \cite{KMNIST} are the data sets with handwritten digits, images of clothing and accessories, and Japanese letters. Apart from MNIST, Semeion \cite{Semeion} is a handwritten digit data set that contains 1593 digits collected from 80 persons. SVHN \cite{SVHN} is another data set for street view house numbers obtained from house numbers in Google Street View images. CIFAR \cite{CIFAR} is a popular data set containing color images. In particular, the CIFAR-10 data set contains 50000 training and 10000 testing images in 10 classes with image resolution of 32x32. CIFAR-100 is similar to the CIFAR-10, except it has 100 classes with 600 images in each class.

Figures \ref{fig:ml_plots_no_data_aug} and \ref{fig:ml_plots_data_aug} show the mean  values and the uncertainty intervals of the training losses for fixed activation function (standard), GAAF, L-LAAF, and N-LAAF, by using the standard deep learning benchmarks. The solid and dashed  lines  are the mean values over three random trials with random seeds. The shaded regions represent the intervals of $2 \times$(the sample standard deviations) for each method. Figures \ref{fig:ml_plots_no_data_aug} and \ref{fig:ml_plots_data_aug} consistently show that adaptive activation accelerates the minimization process of the training loss values. Here, all of  GAAF, L-LAAF and N-LAAF use the slope recovery term, which improved the methods without the recovery term.

The standard cross entropy loss was used for training and plots. We used pre-activation ResNet with  18 layers  \cite{he2016identity} for CIFAR-10, CIFAR-100, and SVHN data sets, whereas we used a standard variant of LeNet \cite{LeCun} with ReLU  for other data sets; i.e., the architecture of the variant of LeNet consists of the following five layers (with the three hidden layers):  (1) input layer, (2) convolutional layer  with 64 $5\times 5$ filters, followed by max pooling of size of 2 by 2 and ReLU, (3) convolutional layer  with 64 $5\times 5$ filters, followed by max pooling of size of 2 by 2 and ReLU, (4) fully connected layer with 1014 output units, followed by ReLU, and (5) Fully connected layer with the number of output units being equal to the number of target classes. All hyper-parameters were fixed a priori across all different data sets and models. We fixed the mini-batch size $s$ to be 64, the initial learning rate to be $0.01$, and the momentum coefficient to be $0.9$. The learning rate was divided by $10$ at the beginning of 10th  epoch for all experiments (with and without data augmentation), and of 100th epoch for those with data augmentation. For the convolutional layers, L-LAAF shares the same single parameter across all pixels and channels within each layer, whereas N-LAAF has as many extra parameters as the number of pixels in every channel. In this section, we used scaling factor $n = 1$ and 2 because the neural network with ReLU activation represents a homogeneous function and hence the behaviors with other values of $n$ can be achieved by changing learning rates in this case. Note that this is not true in previous sections for physics-informed neural network.

\section{Conclusions} 
In summary, we present two versions of locally adaptive activation functions namely, layer-wise and neuron-wise locally adaptive activation functions. Such local activation functions further improve the training speed of the neural network compared to its global predecessor. To further accelerate the training process, an activation slope based \textit{slope recovery} term is added in the loss function for both layer-wise and neuron-wise activation functions, which is shown to enhance the performance of the neural network. To verify our claim, a function approximation problem is solved using deep NN and two inverse PDE problems are solved using physics-informed neural networks, demonstrating that the locally adaptive activations outperform fixed as well as global adaptive activations in terms of training speed and accuracy. Moreover, while the proposed formulations increase the number of additional parameters compared to the fixed activation function, the overall computational cost is comparable. The proposed adaptive activation function with the slope recovery term is also shown to accelerate the training process in standard deep learning benchmark problems. We theoretically prove that no sub-optimal critical point or local minimum attracts gradient descent algorithms in the proposed methods (L-LAAF and N-LAAF) with the slope recovery term under only mild assumptions. We also show that the gradient dynamics of the proposed method is not equivalent to the dynamics of base method with any (adaptive) learning rates. Instead, the proposed methods are equivalent to modifying the gradient dynamics of the base method by implicitly multiplying conditioning matrices to the gradients of the base method. The explicit computations of such matrix-vector products are too expensive for neural networks, whereas our adaptive activation functions efficiently avoid the explicit computations.

\section*{Acknowledgement}
This work was supported by the Department of Energy PhILMs grant DE-SC0019453, and by the DARPA-AIRA grant HR00111990025.

\appendix

\section{Proof of Theorem \ref{thm1}}
We first prove the statement by contradiction for L-LAAF. Suppose that the parameter vector $\hbTheta$ consisting of $\{\mathbf{w}^k,\mathbf{b}^k\}_{k=1}^D$ and $\{a^k\}_{k=1}^{D-1}$ is a limit point of $(\hbTheta_m)_{m\in \NN}$ and  a sub-optimal critical point or a sub-optimal local minimum. 

Let $\ell_{f}^{i}:=\varphi^{i}(u_{\hbTheta}(\rho^i) )$ and $\ell_{u}^i:= |u^{i}-u_{\hbTheta}(\mathbf{x}_u^{i})|^2$. Let $z^{i,k}_{f}$  and $z^{i,k}_{u}$ be the outputs of the $k$-th layer for $\rho^i$ and $(\mathbf{x}_u^{i})$, respectively. Define
  $$
h^{i,k,j}_{f} :=na^{k}(w^{k,j}z^{i,k-1}_f  + b^{k,j})\in \RR,
$$ 
and 
$$
h^{i,k,j}_{u} :=na^{k}(w^{k,j}z^{i,k-1}_u  + b^{k,j})\in \RR, 
$$
for all $j \in \{1,\dots,N_k\}$, where $w^{k,j} \in \RR^{1 \times N_{k-1}}$ and $b^{k,j} \in \RR$. 

Following the proofs in \cite[Propositions 1.2.1-1.2.4]{bertsekas1999nonlinear}, we have that  $\nabla \tJ(\hbTheta)=0$ and $\tJ(\hbTheta)<\tJ c(0)+\mathcal{S}(0)$, for all three cases of the conditions corresponding  the different rules of the learning rate. 
Therefore, 
we have that for  all $k\in \{1,\dots,D-1\},$ 
\begin{align} \label{eq:proof:1}
&\frac{\partial \tJ(\hbTheta)}{\partial{a^{k}}}
\\ \nonumber &=\frac{\partial \mathcal{S}(a)}{\partial{a^{k}}}+\frac{n}{N_{f}}\sum_{i=1}^{N_{f}}  \sum_{j=1}^{N_k}\frac{\partial \ell_{f}^i}{\partial{h^{i,k,j}_{f}}}(w^{k,j}z^{i,k-1}_f  + b^{k,j})+\frac{n}{N_u}\sum_{i=1}^{N_{u}}  \sum_{j=1}^{N_k} \frac{\partial \ell_{u}^i}{\partial{h^{i,k,j}_{u}}}(w^{k,j}z^{i,k-1}_u  + b^{k,j})
\\ \nonumber &= \frac{\partial \mathcal{S}(a)}{\partial{a^{k}}} + \sum_{j=1}^{N_k}\left(\frac{n}{N_{f}}\sum_{i=1}^{N_{f}} \frac{\partial \ell_{f}^i}{\partial{h^{i,k,j}_{f}}}(w^{k,j}z^{i,k-1}_f  + b^{k,j})+\frac{n}{N_u}\sum_{i=1}^{N_{u}} \frac{\partial \ell_{u}^i}{\partial{h^{i,k,j}_{u}}}(w^{k,j}z^{i,k-1}_u  + b^{k,j})\right)
\\ \nonumber & =0.
\end{align}
Furthermore,
 we have that
 for  all $k\in \{1,\dots,D-1\}$ and all $j \in \{1,\dots,N_{k}\}$,
\begin{align} \label{eq:proof:2}
\frac{\partial \tJ(\hbTheta)}{\partial{w^{k,j}}} &=\frac{na^{k}}{N_{f}}\sum_{i=1}^{N_{f}}  \frac{\partial \ell_{f}^i}{\partial{h^{i,k,j}_{f}}}(z^{i,k-1}_f)\T +\frac{na^{k}}{N_u}\sum_{i=1}^{N_{u}}  \frac{\partial \ell_{u}^i}{\partial{h^{i,k,j}_{u}}}(z^{i,k-1}_u)\T ,
\\ \nonumber & =0
\end{align}
and
\begin{align} \label{eq:proof:3}
\frac{\partial \tJ(\hbTheta)}{\partial{ b^{k,j}}}
= \frac{na^{k}}{N_{f}}\sum_{i=1}^{N_{f}}   \frac{\partial \ell_{f}^i}{\partial{h^{i,k,j}_{f}}}+\frac{na^{k}}{N_{f}}\sum_{i=1}^{N_{f}}   \frac{\partial \ell_{u}^i}{\partial{h^{i,k,j}_{u}}}=0.
\end{align}
By combining \eqref{eq:proof:1}--\eqref{eq:proof:3},
for  all $k\in \{1,\dots,D-1\}$,
\begin{align*}
0 & =a^{k}\frac{\partial \tJ(\hbTheta)}{\partial{a^{k}}}
\\ & = a^{k}\frac{\partial \mathcal{S}(a)}{\partial{a^{k}}}+ \sum_{j=1}^{N_k}\left(\frac{na^{k}}{N_{f}}\sum_{i=1}^{N_{f}} \frac{\partial \ell_{f}^i}{\partial{h^{i,k,j}_{f}}}(w^{k,j}z^{i,k-1}_f  + b^{k,j})+\frac{na^{k}}{N_u}\sum_{i=1}^{N_{u}} \frac{\partial \ell_{u}^i}{\partial{h^{i,k,j}_{u}}}(w^{k,j}z^{i,k-1}_u  + b^{k,j})\right)
\\ & = a^{k}\frac{\partial \mathcal{S}(a)}{\partial{a^{k}}}+\sum_{j=1}^{N_k}\left( w^{k,j}\left(\frac{\partial \tJ(\hbTheta)}{\partial{w^{k,j}}}\right)\T +b^{k,j}\left(\frac{\partial \tJ(\hbTheta)}{\partial{ b^{k,j}}} 
\right) \right) = a^{k}\frac{\partial \mathcal{S}(a)}{\partial{a^{k}}}.
\end{align*}
Therefore, 
\begin{align*}
0=a^{k}\frac{\partial \mathcal{S}(a)}{\partial{a^{k'}}}=-a^{k}(D-1)\left(\sum_{k=1}^{D-1} \exp(a^k) \right)^{-2} \exp(a^{k}), 
\end{align*}
which implies that for all  $a^{k}=0$ since $(D-1)\left(\sum_{k=1}^{D-1} \exp(a^k) \right)^{-2} \exp(a^{k})\neq 0$.
This implies that  $\tJ(\hbTheta)=\tJ c(0)+\mathcal{S}(0)$, which contradicts with $\tJ(\hbTheta)<\tJ c(0)+\mathcal{S}(0)$. This proves the desired statement  for L-LAAF. 

For N-LAAF, we  prove the statement by contradiction. Suppose that the parameter vector $\hbTheta$ consisting of $\{\mathbf{w}^k,\mathbf{b}^k\}_{k=1}^D$ and $\{a^k_j\}_{k=1}^{D-1}~~ \forall j = 1,2,\cdots, N_k$ is a limit point of $(\hbTheta_m)_{m\in \NN}$ and  a sub-optimal critical point or a sub-optimal local minimum. Redefine 
  $$
h^{i,k,j}_{f} :=na^{k}_{j}(w^{k,j}z^{i,k-1}_f  + b^{k,j})\in \RR,
$$ 
and 
$$
h^{i,k,j}_{u} :=na^{k}_{j}(w^{k,j}z^{i,k-1}_u  + b^{k,j})\in \RR, 
$$
for all $j \in \{1,\dots,N_k\}$, where $w^{k,j} \in \RR^{1 \times N_{k-1}}$ and $b^{k,j} \in \RR$. Then, by the same proof steps, we have that  $\nabla \tJ(\hbTheta)=0$ and $\tJ(\hbTheta)<\tJ c(0)+\mathcal{S}(0)$, for all three cases of the conditions corresponding  the different rules of the learning rate. Therefore, 
we have that for  all $k\in \{1,\dots,D-1\}$ and all $j \in \{1,\dots,N_{k}\}$, 
\begin{align} \label{eq:proof:4}
\frac{\partial \tJ(\hbTheta)}{\partial{a^{k}_j}} &=\frac{n}{N_{f}}\sum_{i=1}^{N_{f}}  \frac{\partial \ell_{f}^i}{\partial{h^{i,k,j}_{f}}}(w^{k,j}z^{i,k-1}_f  + b^{k,j})+\frac{n}{N_u}\sum_{i=1}^{N_{u}}  \frac{\partial \ell_{u}^i}{\partial{h^{i,k,j}_{u}}}(w^{k,j}z^{i,k-1}_u  + b^{k,j})+\frac{\partial \mathcal{S}(a)}{\partial{a^{k}_j}}
\\ \nonumber & =0.
\end{align}
By combining \eqref{eq:proof:2}--\eqref{eq:proof:4},
for  all $k\in \{1,\dots,D-1\}$ and all $j \in \{1,\dots,N_{k}\}$, ,
\begin{align*}
0 & =a^{k}_j\frac{\partial \tJ(\hbTheta)}{\partial{a^{k}_j}}
\\ & =\frac{na_{j}^{k}}{N_{f}}\sum_{i=1}^{N_{f}} \frac{\partial \ell_{f}^i}{\partial{h^{i,k,j}_{f}}}(w^{k,j}z^{i,k-1}_f  + b^{k,j})+\frac{na_{j}^{k}}{N_u}\sum_{i=1}^{N_{u}} \frac{\partial \ell_{u}^i}{\partial{h^{i,k,j}_{u}}}(w^{k,j}z^{i,k-1}_u  + b^{k,j})+a_{j}^{k}\frac{\partial \mathcal{S}(a)}{\partial{a_{j}^{k}}}
\\ & = w^{k,j}\left(\frac{\partial \tJ(\hbTheta)}{\partial{w^{k,j}}}\right)\T +b^{k,j}\left(\frac{\partial \tJ(\hbTheta)}{\partial{ b^{k,j}}} 
\right) +a_{j}^{k}\frac{\partial \mathcal{S}(a)}{\partial{a_{j}^{k}}} =a_{j}^{k}\frac{\partial \mathcal{S}(a)}{\partial{a_{j}^{k}}}.
\end{align*}
Therefore, 
\begin{align*}
0=a_{j}^{k}\frac{\partial \mathcal{S}(a)}{\partial{a^{k'}}}=-2a_{j}^{k}(D-1)\left(\sum_{k=1}^{D-1} \exp\left(\frac{\sum_{i=1}^{N_k} a_i^k}{N_k} \right) \right)^{-2} \exp\left(\frac{\sum_{i=1}^{N_k} a_i^k}{N_k}\right)/N_k, 
\end{align*}
which implies that for all  $a_j^{k}=0$ since $(D-1)\left(\sum_{k=1}^{D-1} \exp\left(\frac{\sum_{i=1}^{N_k} a_i^k}{N_k} \right) \right)^{-2} \exp\left(\frac{\sum_{i=1}^{N_k} a_i^k}{N_k}\right)\neq 0$.
This implies that  $\tJ(\hbTheta)=\tJ c(0)+\mathcal{S}(0)$, which contradicts with $\tJ(\hbTheta)<\tJ c(0)+\mathcal{S}(0)$. This proves the desired statement  for N-LAAF.

\qed

\end{document}